\documentclass[lettersize,journal, compsoc]{IEEEtran} %
\usepackage{amsmath,amsfonts}
\usepackage{algorithmic}
\usepackage{algorithm}
\usepackage{array}
\usepackage{textcomp}
\usepackage{stfloats}
\usepackage{url}
\usepackage{verbatim}
\usepackage{graphicx}
\usepackage{subfigure}
\usepackage{cite}
\usepackage{color}
\usepackage{multirow}
\usepackage{enumitem}
\usepackage{todonotes}
\setlist[enumerate]{leftmargin=*}

\usepackage[utf8]{inputenc} 
\usepackage[T1]{fontenc}    
\usepackage{hyperref}       
\usepackage{url}            
\usepackage{booktabs}       
\usepackage{amsfonts}       
\usepackage{nicefrac}       
\usepackage{microtype}      
\usepackage{xcolor}         
\usepackage{ulem}
\usepackage{dsfont}
\usepackage[font=footnotesize]{caption}
\newcommand{\etal}{\textit{et~al.}}
\newcommand\ie{\textit{i.e.}}
\newcommand\eg{\textit{e.g.}}

\begin{document}

\title{DreamStory: Open-Domain Story Visualization by LLM-Guided Multi-Subject Consistent Diffusion}

\author{
Huiguo He, Huan Yang, Zixi Tuo, Yuan Zhou, Qiuyue Wang, Yuhang Zhang, Zeyu Liu, 
\\Wenhao Huang, Hongyang Chao, Jian Yin
    \IEEEcompsocitemizethanks{\IEEEcompsocthanksitem Huiguo He (hehg3@mail2.sysu.edu.cn) and Hongyang Chao are with the School of Computer Science and Engineering, Sun Yat-sen University, Guangzhou, Guangdong, China. \protect\\
    \vspace{-8pt}
    \IEEEcompsocthanksitem Jian Yin is with the School of Artificial Intelligence, Sun Yat-sen University, Zhuhai, Guangdong, China.\protect\\
    \vspace{-8pt}
    \IEEEcompsocthanksitem Huan Yang (hyang@fastmail.com), Zixi Tuo, Yuan Zhou, Qiuyue Wang, Yuhang Zhang, Zeyu Liu, and Wenhao Huang are with 01.AI, Beijing, Beijing, China. \protect\\
    \vspace{-8pt}
    \IEEEcompsocthanksitem This work was done when Huiguo He was an intern at 01.AI.
    }
    \thanks{Huan Yang and Jian Yin are the corresponding authors.}
}

\markboth{Journal of \LaTeX\ Class Files,~Vol.~14, No.~8, August~2021}%
{Shell \MakeLowercase{\textit{et al.}}: A Sample Article Using IEEEtran.cls for IEEE Journals}

\IEEEpubid{0000--0000/00\$00.00~\copyright~2021 IEEE}

\IEEEtitleabstractindextext{%
\begin{center} \setcounter{figure}{0}
    \setlength{\abovecaptionskip}{-0.1cm} 
    \setlength{\belowcaptionskip}{-0.4cm} 
    \includegraphics[width=0.98\linewidth]{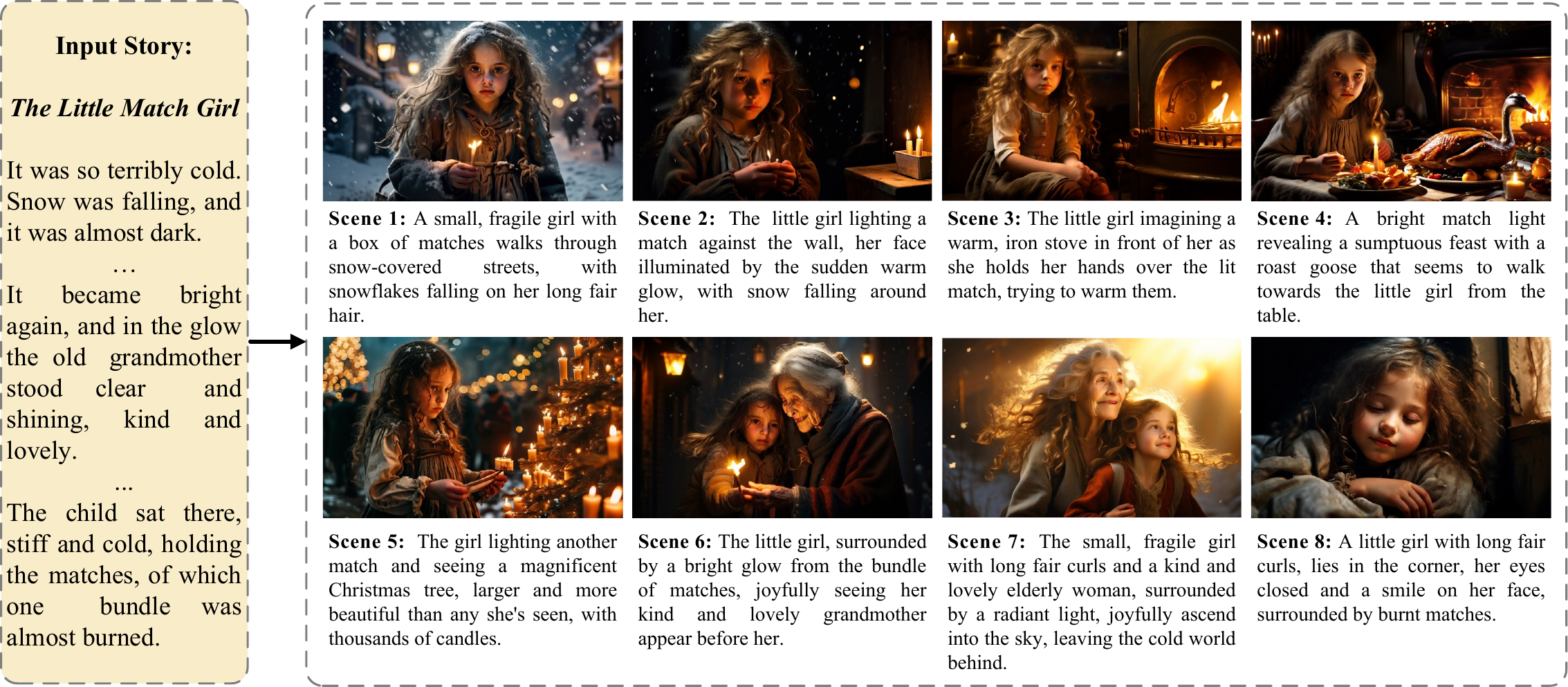}
     \begin{flushleft}
        \footnotesize
        \captionof{figure}{Illustration of our proposed DreamStory framework. This system takes a full narrative text as input, generates vivid visual content, and maintains the consistency of multiple subjects across various scenes within the story.
        Please visit the project \href{https://dream-xyz.github.io/dreamstory}{homepage} to watch the video.
        } 
     \end{flushleft}
     \vspace{-2pt}
    \label{fig:teaser}
    \vspace{-2pt}
\end{center}

\begin{abstract}
Story visualization aims to create visually compelling images or videos corresponding to textual narratives. Despite recent advances in diffusion models yielding promising results, existing methods still struggle to create a coherent sequence of subject-consistent frames based solely on a story. 
To this end, we propose \textbf{DreamStory}, an automatic open-domain story visualization framework by leveraging the LLMs and a novel multi-subject consistent diffusion model. 
DreamStory consists of (1) an LLM acting as a story director and (2) an innovative \textbf{M}ulti-\textbf{S}ubject consistent \textbf{D}iffusion model (\textbf{MSD}) for generating consistent multi-subject across the images. 
First, DreamStory employs the LLM to generate descriptive prompts for subjects and scenes aligned with the story, annotating each scene's subjects for subsequent subject-consistent generation. Second, DreamStory utilizes these detailed subject descriptions to create portraits of the subjects, with these portraits and their corresponding textual information serving as multimodal anchors (guidance). Finally, the MSD uses these multimodal anchors to generate story scenes with consistent multi-subject. 
Specifically, the MSD includes Masked Mutual Self-Attention (MMSA) and Masked Mutual Cross-Attention (MMCA) modules. MMSA module ensures detailed appearance consistency with reference images, while MMCA captures key attributes of subjects from their reference text to ensure semantic consistency. Both modules employ masking mechanisms to restrict each scene's subjects to referencing the multimodal information of the corresponding subject, effectively preventing blending between multiple subjects.
To validate our approach and promote progress in story visualization, we established a benchmark, \textbf{DS-500}, which can assess the overall performance of the story visualization framework, subject-identification accuracy, and the consistency of the generation model. 
Extensive experiments validate the effectiveness of DreamStory in both subjective and objective evaluations.
Please visit our project homepage at \href{https://dream-xyz.github.io/dreamstory}{https://dream-xyz.github.io/dreamstory}.
\end{abstract}

\vspace{-6pt}

}

\maketitle

\section{Introduction}\label{sec:intro}

Story visualization aims to create a visually captivating and coherent sequence of visual content (including images and videos) that aligns with a given story. This field has become increasingly important in entertainment~\cite{SVentertainment1} and education~\cite{SVeducation1, SVeducation2}. However, story visualization is particularly daunting in open-domain contexts, subjectized by diverse content and themes. Despite the significant advancements in diffusion models~\cite{song2020DDIM_diffusion, song2020score_diffusion, sd_ldm_diffusion, ramesh2022hierarchical_DALLE2, podell2023sdxl, li2024playground, TPAMIcreateYourWorld, TPAMIdiffusionsurvey, TPAMIDreamStone, TPAMIImageSynthesis, TPAMIMotionDiffuse} for visual content creation, current methods still struggle with the challenge of directly translating textual narratives into corresponding consistent visual representations.

The primary challenges stem from two key issues. The first is generating effective story prompts using a Large Language Model (LLM) while accurately identifying recurring subjects within the narrative. The second is seamlessly incorporating this information into the diffusion model to maintain consistency in multi-subject generation has proven to be a significant obstacle. On the one hand, LLM has recently demonstrated impressive capabilities in long-text understanding~\cite{chen2023longlora, meister2021sparse} and In-Context Learning (ICL)~\cite{llm_survey}. Chain-of-Thought (CoT)~\cite{wei2022chain, chung2024scaling, yao2024tree} reasoning has significantly improved their performance in handling intricate text understanding tasks. However, in the context of story visualization, the effectiveness of LLMs heavily depends on selecting appropriate prompts, posing challenges for generating coherent visual content and requiring further refinement.

On the other hand, although previous works~\cite{avrahami2023chosen,ye2023_ip_adapter, ruiz2023dreambooth, MUDI, storygen_liu2023intelligent, cao2023masactrl, tewel2024_ConsiStory} have made efforts to improve the consistency in the diffusion model, their attempts have not yielded a satisfactory level of multi-subject consistency in open-domain stories. These methods fall into four categories: (1) dataset-based training, (2) few-shot fine-tuning, (3) encoder-based, and (4) training-free methods. 
Dataset-based methods~\cite{storygen_liu2023intelligent} rely on specific story datasets (\eg, PororoSV~\cite{StoryGAN_2019_CVPR} and FlintstonesSV~\cite{FlintstonesSV}). Therefore, they are closed-domain methods and are limited in open-domain capabilities. 
Few-shot fine-tuning methods~\cite{ruiz2023dreambooth, MUDI} offer customization but necessitate additional training costs for each story, inevitably leading to overfitting and diversity degradation.
Encoder-based methods~\cite{ye2023_ip_adapter, controlnet} aims to train an image encoder to convert the reference image into the image condition aligned with the original text condition. It guides the generation process by injecting image conditions into the cross-attention layer. However, these methods mainly focus on a single subject and are hindered by diversity degradation~\cite{tewel2024_ConsiStory} and computational resource consumption.
Training-free methods~\cite{cao2023masactrl, tewel2024_ConsiStory, storydiffusion} maintain subject consistency by facilitating interaction between target and reference images in the self-attention layer. These methods have gained widespread attention due to their efficiency. However, they still face issues of subject confusion and overlooking fine-grained descriptions in open-domain story visualization.

To address the aforementioned challenges, we introduce \textbf{DreamStory}, a training-free, automatic, open-domain story visualization framework. Specifically, given a story, DreamStory first employs the LLM (such as GPT-4~\cite{openai2023GPT4} and Yi~\cite{01ai_young2024yi}) to generate detailed descriptive prompts for both subjects and scenes, ensuring alignment with the narrative. This includes annotating each scene's subjects, as well as performing necessary rewrites, for subsequent consistent generation. 
Subsequently, DreamStory utilizes these detailed subject descriptions to create accurate portraits of the subjects. These portraits, along with their corresponding text, are served as multimodal anchors (guidance) for the consequent generation process. The motivation of this approach is that the text and images are naturally aligned in the diffusion model's semantic space, as the image is generated based on the corresponding text. Therefore, this aligned multimodal information, which is rich in subjects' semantics, attributes, and visual appearance, benefits the model in generating more consistent subjects.

Finally, a novel \textbf{M}ulti-\textbf{S}ubject consistent \textbf{D}iffusion model (\textbf{MSD}) utilizes these multimodal anchors to produce story scenes that maintain consistency across multiple subjects. 
The MSD consists of two key modules: Masked Mutual Self-Attention (MMSA) and Masked Mutual Cross-Attention (MMCA). The MMCA module captures essential subject attributes to ensure semantic consistency. While previous studies show that aligning image encoders with text embeddings maintains layout~\cite{controlnet} and identity~\cite{ye2023_ip_adapter} consistency, our MMCA module uniquely preserves subject attributes (\eg, clothing and accessories) by using the naturally aligned text.
In parallel, the MMSA module maintains detailed appearance consistency by allowing the Query ($Q$) token to query the Key ($K$) and Value ($V$) tokens that belong to the same subject in the anchor. 
Our specific innovation is characterized by the use of masking mechanisms in both modules, ensuring that each scene’s subjects only reference information pertinent to the corresponding subject. This approach effectively prevents the blending of attributes between different subjects, thereby preserving their individual consistency.

To validate our approach and promote progress in story visualization, we established a benchmark, \textbf{DS-500}, comprising 100 stories and 400 synthetic samples. The
100 Stories assesses the overall framework of automatic open-domain story visualization. The remaining 400 synthetic samples, each with 0, 1, 2, and 3 subjects, are utilized to evaluate the precision of the LLM in annotating scene subjects and the consistency of multi-subject generation.

The main contributions of this paper are as follows:
\begin{enumerate}
    \item We introduce \textbf{DreamStory},  a training-free framework for automatic open-domain story visualization, which utilizes LLMs as a story director to generate concise prompts of subjects and scenes, annotating the subjects in each scene. This information guides diffusion models in creating visually consistent content that aligns with the story narrative.
    \item We propose a novel Multi-Subject consistent Diffusion model (\textbf{MSD}) that leverages both the subject prompt and its corresponding portrait to maintain consistency in multiple subjects across frames.
    \item We build an evaluation benchmark \textbf{DS-500} for open domain story visualization. Our method outperforms the mainstream methods on aesthetics, image-text consistency, and subject consistency through objective and subjective evaluations.
\end{enumerate}


\section{Related Works}\label{sec:related_works}

\subsection{Visual Content Generation}\label{subsec:related_works_DM}

Variational AutoEncoders (VAEs)~\cite{VAE} and Generative Adversarial Networks (GANs)~\cite{StoryGAN_2019_CVPR,song2020character_GAN, li2022clustering_GAN, ma2022ai} used to dominate visual generation field. Despite the significant advancements made by GAN, its optimization challenges persist~\cite{metz2016unrolledGAN, arjovsky2017WGAN, gulrajani2017improvedWGAN}. Later, diffusion-based generative models~\cite{song2020DDIM_diffusion, DDPM_diffusion, song2020score_diffusion, nichol2021improved_diffusion, dhariwal2021_diffusion, saharia2022palette} have emerged, achieving impressive image quality and diversity. 
Notably, Stable Diffusion (SD)~\cite{sd_ldm_diffusion} utilizes a diffusion model in latent space, trained on the largest LAION-5B~\cite{schuhmann2022laion_5B} dataset. 
While subsequent studies~\cite{podell2023sdxl, chen2023pixart, li2024playground} have improved resolution and aesthetics, ensuring subject consistency across multiple images remains a challenge.

\subsubsection{Dataset-based Story Visualization}

Early methods~\cite{SV_GAN1, SV_GAN2, SV_T1, CVPR2023_Make_a_Story, gu2023tevis_retrieval, storydalle, ARLDM} for story visualization relied on collecting datasets, such as PororoSV~\cite{StoryGAN_2019_CVPR} and FlintstonesSV~\cite{FlintstonesSV}. For example, Rahman~\etal~\cite{CVPR2023_Make_a_Story} propose a novel autoregressive
diffusion-based framework. This framework includes a visual memory module that implicitly captures the actor and background context across the generated frames. Pan~\etal~\cite{ARLDM} propose an auto-regressively diffusion model conditioned on history captions and generated images. It employs multimodal guidance (a CLIP~\cite{radford2021CLIP} text encoder and a BLIP~\cite{BLIP, BLIP2} multimodal encoder) to ensure the generation of relevant and coherent images. Liu~\etal~\cite{storygen_liu2023intelligent} further proposed the StorySalon dataset and achieved SOTA results.

However, these methods are constrained by the size and quality of existing datasets, limiting their performance in open-domain tasks. In contrast, our approach is designed for open-domain scenarios and is training-free, circumventing the challenges of gathering high-quality story visualization datasets.

\subsubsection{Few-shot Finetuning Consistent Generation}

Few-shot finetuning methods~\cite{ruiz2023dreambooth, avrahami2023chosen, PortraitBooth, TPAMIcreateYourWorld, MUDI, multibooth, DCO} primarily revolve around personalized image generation based on a few subject images. The model is finetuned on these images to learn their unique textual expressions. For example, Dreambooth~\cite{ruiz2023dreambooth} first proposed fine-tuning SD with LORA~\cite{hu2021lora} on several images to make the model remember specific subject tokens for reference images. Sun~\etal~\cite{TPAMIcreateYourWorld} further extend it in a never-ending manner, \ie, new concepts from the user are quickly learned without catastrophic forgetting. Jang~\etal~\cite{MUDI} proposed using a segmentation model to segment subjects for training and inference,  effectively mitigating the influence of multi-subject blending. It has achieved SOTA performance in the field of few-shot finetuning for multi-subject consistent generation.

However, these methods necessitate finetuning for each story or subject, resulting in extra computational costs. Besides, this approach inevitably risks overfitting, leading to a decline in the aesthetic quality and diversity of the generated images~\cite{tewel2024_ConsiStory}.

\subsubsection{Encoder-Based Consistent Generation}

In foundational T2I models, such as SD~\cite{sd_ldm_diffusion} and SDXL~\cite{podell2023sdxl}, the text is typically encoded into an embedding vector and injected into a cross-attention mechanism to generate images satisfying textual conditions. To achieve consistent generation, previous methods~\cite{arar2023domain, IDAdapter, AnyDoor, PhotoMaker, ye2023_ip_adapter, BLIPdiffusion} attempt to design an image encoder for generation under the image condition. Specifically, some studies~\cite{IDAdapter, PhotoMaker} tried to train a face encoder to ensure that the generated images maintain ID consistency. Similarly, Ye~\etal~\cite{ye2023_ip_adapter} tried to train an image encoder that converts image conditions into a space aligned with the original text embedding. However, these methods can only handle a single subject. Therefore, they are unsuitable for open-domain story visualization, which may involve multiple subjects of various types, including anthropomorphized animals. 

\subsubsection{Training-free Consistent Generation}

Training-free methods have gained widespread attention due to their efficiency. These methods~\cite{cao2023masactrl,tewel2024_ConsiStory,storydiffusion} maintain subject consistency by facilitating interaction between the target and reference images in the self-attention layer. For example, MasaCtrl~\cite{cao2023masactrl} introduced mutual self-attention, which replaces the \textit{key} and \textit{value} in self-attention with those from the reference image. They also utilized a cross-attention map as a mask to ensure that mutual self-attention concentrates on relevant subjects. ConsiStory~\cite{tewel2024_ConsiStory} introduced Subject Driven Self-Attention (SDSA), which allows each frame to refer all subjects from multiple reference images in a batch. They also implemented token dropout and blended Vanilla Query techniques to increase layout diversity, and used DIFT~\cite{DIFT} for feature injection in self-attention to enhance detail consistency. 

However, these methods still struggle to generate multiple subjects because all subjects in the target image can refer to all reference images regardless of whether their roles are the same. Furthermore, they failed to consider fine-grained descriptions of subjects that contain rich information on attributes which are beneficial for maintaining consistency.

\subsection{Large Language Model}\label{subsec:related_works_LLM}

\subsubsection{LLM in Text Understanding}

Large Language Model (LLM) has recently demonstrated impressive capabilities in various NLP tasks, such as text summarization~\cite{cao2018faithful_text_sum, gui2019attention_text_sum}, and question answering~\cite{wang2018r_QA, min2018efficient_QA}. 
Moreover, ChatGPT employs Reinforcement Learning from Human Feedback (RLHF)~\cite{RLHF_ouyang2022training} to align the model’s output with human preferences, demonstrating an impressive ability for human interaction. Its remarkable In-Context Learning (ICL)~\cite{llm_survey} ability enables it to generate expected outputs by completing the input text’s word sequence, without additional fine-tuning. Furthermore, some studies~\cite{wei2022chain, chung2024scaling, yao2024tree} have revealed that carefully crafted Chain of Thought (CoT) strategies can significantly improve the performance of LLM models in handling intricate and lengthy tasks.
Though LLM models can summarize texts and answer human questions, their ability to generate suitable prompts that guide diffusion models for story visualization is less studied. In the story visualization field, diffusion models are limited to recognizing subjects in novel visual vocabulary as they are usually referred to by their names without visual descriptions. In this paper, we study how to adjust the prompts generated by LLM models to better bootstrap diffusion models for story visualization.

\subsubsection{LLM in Image Generation }
Advanced visual generation models struggle with low-quality descriptions, which impedes their comprehension of subtle semantics. Many research~\cite{segalis2023picture, wen2023improving, blattmann2023stable, hao2022promptist_furu, MVP, autostudio} efforts have aimed to enhance the capabilities of T2I models by refining datasets and modifying user prompts. Specifically, Segalis~\etal~\cite{segalis2023picture} enhanced generation performance by re-captioning the images using a specialized LLM and retraining a text-to-image model on the updated data.
Some approaches propose rewriting user prompts to enhance generated images regarding the aesthetic~\cite{hao2022promptist_furu} and NFT market values~\cite{MVP}. Yang~\etal~\cite{RPG} utilizes language models for planning, recaptioning, and generating images with coherent layouts. Cheng~\etal~\cite{theatergen} proposes employing LLM to facilitate user editing in an interactive manner. Zhuang~\etal~\cite{4vlogger} leverages an LLM-guided pipeline to generate minute-long vlogs with coherent storylines and diverse scenes.

These methods all indicate that a powerful LLM has the potential to enhance image generation. In this article, we utilize the LLM as a director to guide the generation of a series of story images with consistent multi-subject.

\begin{figure*}[t]
\setlength{\abovecaptionskip}{2pt} 
\setlength{\belowcaptionskip}{-0.4cm} 
\centering
\includegraphics[width=1.0\linewidth]{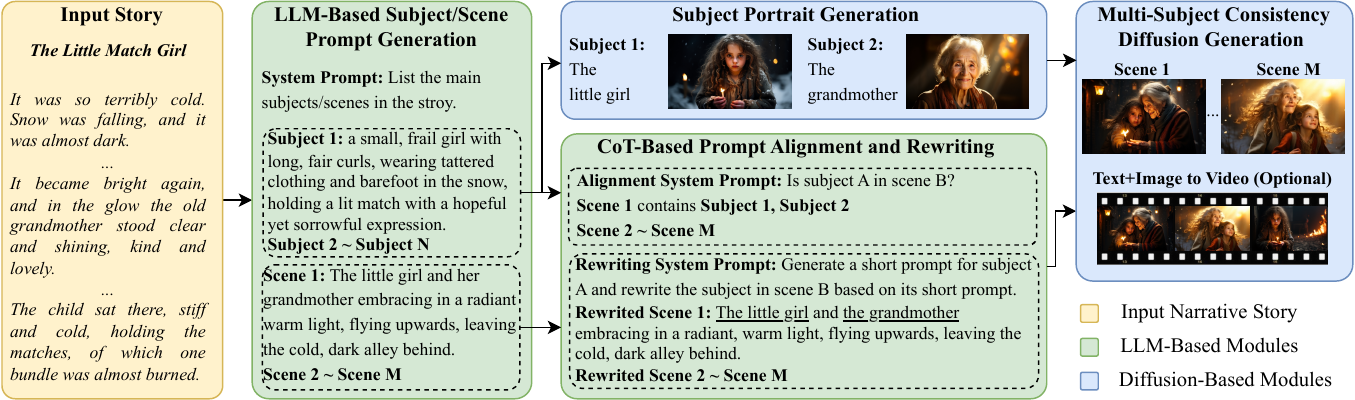}
\caption{The framework of our proposed DreamStory.
    Initially, the LLM comprehends a story and generates detailed prompts for key subjects and scenes. 
    These prompts are aligned and rewritten to enhance understanding of the diffusion model, ensuring accurate visual content generation. 
    Subject portraits are then generated based on these prompts, serving as multimodal anchors for maintaining multi-subject consistency and enriching scenes with high-quality visual details, which facilitates subsequent video creation using an image-to-video model.}
\label{fig:framework}

\end{figure*}

\section{Our Approach}\label{sec:method}

\subsection{Overall Framework}\label{subsec:method_auto_AI_pipeline}

In this subsection, we introduce our automated story visualization framework (DreamStory), shown in Fig.~\ref{fig:framework}. The framework operates as follows:

\begin{enumerate}
    \renewcommand{\labelenumi}{\theenumi)}
    \item \textbf{Story comprehension and prompt generation}. Given a story (\eg, \textit{The Little Match Girl}), a Large Language Model (LLM), such as GPT-4~\cite{openai2023GPT4}, comprehends the narrative and generates concise yet detailed prompts for key subjects and scenes. These prompts serve as the foundation for subsequent visual content generation.
    \item \textbf{Prompt alignment and rewriting}. The LLM identifies the subjects within each scene and performs necessary rewrites, replacing names with descriptions that the diffusion model can understand, such as rewriting "Kondo" to "towering gorilla." This enriches the scenes for visual content generation.
    \item \textbf{Subject portraits generation}. The Text-to-Image (T2I) model then utilizes these prompts to create subject portraits. By focusing on individual subjects, this approach ensures alignment with the provided prompts.
    \item \textbf{Multimodal anchors for scene generation}. The subject portraits, accompanied by their textual descriptions, act as multimodal anchors. The subsequent T2I model leverages these multimodal anchors to maintain subject consistency. It enriches the scenes with additional details, resulting in high-quality visual representations. These images can be transformed into video clips using an Image-to-Video (I2V) model, such as SVD~\cite{blattmann2023stable}, ConsistI2V~\cite{consisti2v}, and Kling~\footnote{\href{https://kling.kuaishou.com/en}{Kling}}, or extended into longer videos with models like SEINE~\cite{seine}.
\end{enumerate}
Our comprehensive process enhances the final image quality, making DreamStory indispensable for vivid story visualization.

\subsection{LLM Prompt Generation Model}\label{subsec:method_LLM_prompt_gen}
The Chain of Thought (CoT)~\cite{wei2022chain_COT1, wang2023towards_COT2, feng2024towards_COT3, chu2023survey_COT} strategy has shown promising results in LLMs. The core idea of CoT is to break down complex problems into a series of simpler, manageable tasks, which guides the model towards generating anticipated results and enhances overall performance~\cite{wei2022chain_COT1, wang2023towards_COT2}.

Inspired by these pioneering works, we designed a prompt generation model based on the CoT strategy for the diffusion model. Our approach simplifies the entire process into a sequence of simple steps: generating prompts for subjects or scenes, annotating whether subjects are present in scenes, and making necessary revisions. Each of these tasks (text understanding or rewriting) is considerably easier due to its widespread presence in the LLM’s training samples compared with that of directly obtaining a suitable prompt for the diffusion model to visualize stories. All the task prompts are designed with at least two in-context examples to improve the performance and formatting of the results~\cite{chu2023survey_COT}.

In the process of annotating scenes, we utilize the LLM to determine if a subject is present in the scene’s imagery, given the subject’s name and detailed prompts. We have observed that the LLM often generates scene prompts using the subject’s name, such as "Kondo". However, these prompts encounter difficulties when applied to the diffusion model, which often fails to recognize the names of subjects, particularly when the subject is not well-known and is absent from the training data.

To address this issue, we propose rewriting the scene prompts. Specifically, we employ the LLM to create a concise prompt for the subject that encapsulates its key attributes. We then instruct the LLM to rewrite the scene based on this newly created short prompt. For example, the subject “Kondo” would be replaced with a description such as "towering gorilla". This method ensures a more accurate visual representation of the subject within the scene and is more suitable for the diffusion model.

Our approach to LLM prompt generation presents a logical sequence of steps that address the challenges of generating detailed descriptions in vivid stories, as highlighted in Section~\ref{sec:intro}. It provides a structured way to generate appropriate prompts with precise details for the diffusion model. It should be noted that our approach can generate an arbitrary number of scenes, which can be specified by the user or determined by the LLM based on the story content.

\begin{figure*}[t]
\setlength{\abovecaptionskip}{2pt} 
\setlength{\belowcaptionskip}{-0.2cm} 
\centering
\includegraphics[width=1.0\linewidth]{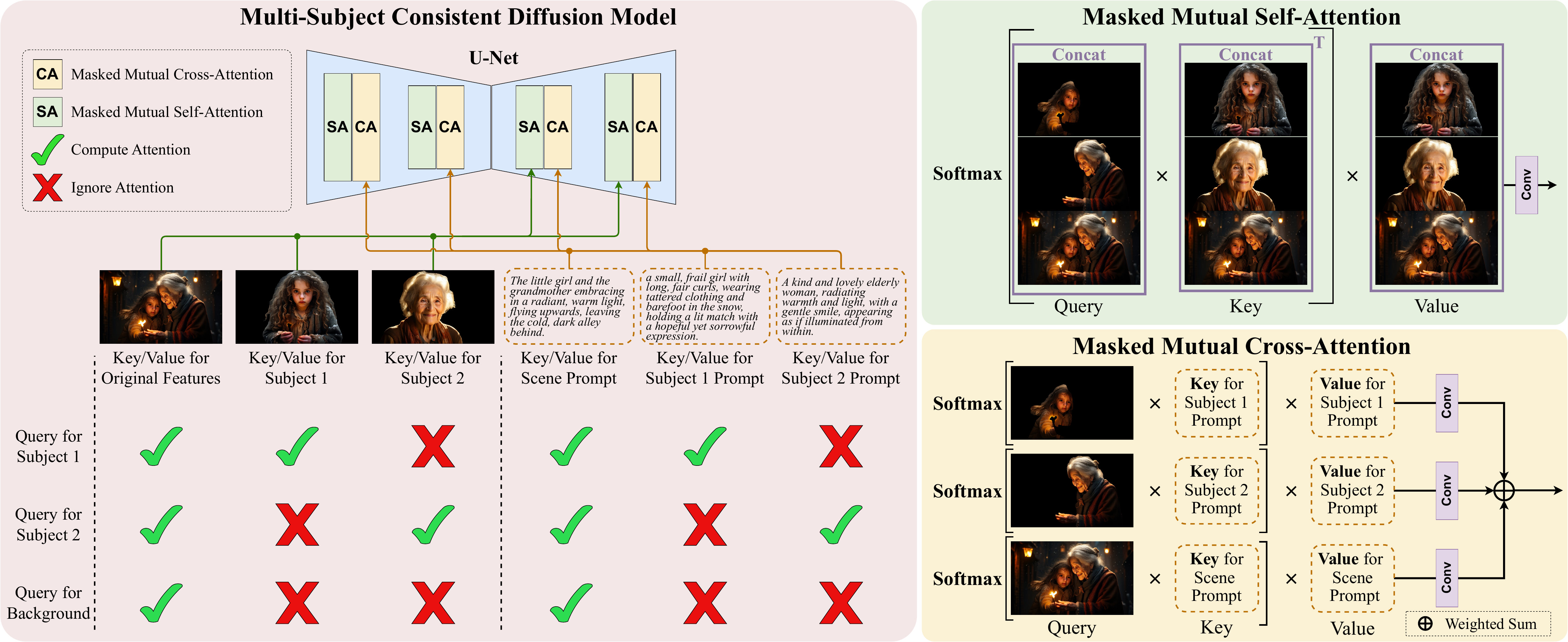}
\caption{The illustration of our Multi-Subject consistent Diffusion models (MSD), along with its Masked Mutual Self-Attention (MMSA) and Masked Mutual Cross-Attention (MMCA) mechanisms. It uses two subjects as examples and can be extended to any number of subjects.
Query, Key, and Value projections in the attention layer have been omitted for ease of presentation.}
\label{fig:attention}

\end{figure*}

\subsection{Multi-Subject Consistent Diffusion Model}\label{subsec:method_consis_T2I}

Preserving subject consistency is a crucial objective in the generation of story images. Our MSD is specifically designed to provide a training-free solution for open-domain story visualization, as shown in Fig.~\ref{fig:attention}. This approach is necessitated by the considerable costs involved in obtaining high-quality datasets for story visualization.

\subsubsection{Existing Attention Mechanism}
Assume the resolution of the generated image is downsampled to $H \times W$ in the attention layer, and $B$ and $C$ respectively denote the batch size and the number of channels.
A standard attention layer in the popular diffusion model (\eg, SD~\cite{sd_ldm_diffusion}, SD-XL~\cite{podell2023sdxl}, and Playground~\cite{li2024playground}) can be formulated as follows, 
\begin{align}
\label{eq_softmax}
    A_i &= \textit{softmax}\left({Q_i K_i}/{\sqrt{d_k}}\right), \\
    O_i &= \textit{conv}_{\scriptsize \tiny out}(A_i \cdot V_i), 
\end{align}
where the $O_i \in \mathbb{R}^{\footnotesize{B \times C \times HW}}$ represents the attention output and $A_i \in \mathbb{R}^{\footnotesize{BC \times HW \times S}}$ indicates the attention weight for the \textit{i}-th image, with $S$ being the embedding length of key-value features.
For self-attention, the key-value features $KV$ come from spatial features with embedding length $S=HW$, while for cross-attention, they are derived from text embeddings where $S=T=77$ in U-Net-based models~\cite{sd_ldm_diffusion,podell2023sdxl,li2024playground}. The query features $Q$ are always projected from spatial features through learned projection matrices.
A simple convolution layer $\textit{conv}_{\scriptsize \tiny out}$ is finally applied to fuse the output features. We omitted the residual connection and layer count to simplify our expression. 

Existing work has verified that self-attention can control the appearance of the generated image~\cite{alaluf2023cross_appearance}, while cross-attention controls the layout and can be used to locate the area of the target subject~\cite{chefer2023attend}. 
Based on these discoveries, recent works have verified that the appearance information of reference images can be injected into the generation process by substiting~\cite{cao2023masactrl} or cascading~\cite{tewel2024_ConsiStory} $K$ and $V$ with that of in the reference images. In the multi-subject scenario, however, they failed to keep multi-subject consistency because all subjects in the target image can refer to the information from all reference images regardless of whether their roles are the same.

\subsubsection{Accurate Object Mask Generation} 

Accurate subject mask generation has been verified as a crucial problem in image generation~\cite{cao2023masactrl, tewel2024_ConsiStory} and editing~\cite{BLIPdiffusion, SCFG}. However, obtaining the subject mask in an unregenerated target image is difficult. Previous works utilize LLM to manage the layout of generated images for editing~\cite{theatergen} and accurate attribute binding~\cite{RPG}. This potentially leads to a lack of aesthetic layout and the generation of objects with unreasonable sizes, as LLM has not been optimized in this situation.
Since diffusion models tend to generate similar layouts under close control conditions if the random seeds are the same~\cite{layoutWACV}, we adopt an open-vocabulary segmentation model, \eg, GroundingSAM~\cite{GroundedSAM}, to obtain an accurate subject mask in rehearsal target images, which is pre-generated with the original diffusion model. The detection phrases are marked by LLM, as mentioned above. To improve accuracy, we contact tokens of all subjects separated by periods as detection prompts for the target image, such as "man. girl.". A simple post-processing is adopted to guarantee the non-overlapping of all masks, enhancing the robustness of our approach.

Due to the imposition of a new control process, the image may go beyond the SAM's mask during generation, especially in the later steps. So, it is necessary to adjust the mask according to the features in the generation process. Previous work~\cite{cao2023masactrl, tewel2024_ConsiStory} mainly adopts a segmentation mask by averaging the cross-attention maps of the subject token. However, this strategy may create a hole and a noisy mask~\cite{SCFG}. Therefore, we obtain a segmentation mask by multiplying the self-attention and cross-attention maps, with the self-attention map serving as a completion of the cross-attention maps~\cite{SCFG}. So, the semantics maps are calculated as follows,
\begin{align}
    m_\text{\scriptsize \tiny TGT}^{i} &= \textit{mean}( \sum_{l\in L} \sum_{r\in R} (A_{sa})^r \times A_{ca} ), \\
    M_{i} &= \textit{Otsu}\Big( m_\text{\scriptsize \tiny TGT}^{i}\Big) \times \textit{Otsu}\Big(m_\text{\scriptsize \tiny REF}^{i} \Big)^{\top}, 
\end{align}
where $A_{sa} \in \mathbb{R}^{\footnotesize{BC \times HW \times HW}}$ and $A_{ca} \in \mathbb{R}^{\footnotesize{BC \times HW \times T}}$ denote the self-attention and cross-attention layers of the target image in the same block. 
The flattened mask of the reference image and target image $m^i \in \mathbb{R}^{\footnotesize{B \times HW}}$ are denoted with the subscript ‘\textit{REF}’ and ‘\textit{TGT}’, respectively.
$L$ denotes the layer for gathering the attention map, and the averaging operation is denoted by $\textit{mean}(\cdot)$. $R$ is a hyper-parameter set as 4 followed by~\cite{SCFG}. To save computational cost, we only collect all layers $L$ from the previous timestep $t-1$ to calculate the mask.
The threshold is determined by Otsu's method~\cite{otsu1975threshold}, represented as $\textit{Otsu}(\cdot)$. The symbol $\times$ stands for matrix multiplication. The final matrix $M_{i} \in \mathbb{R}^{\footnotesize{B \times HW \times HW}}$ illustrates the correlation between the elements of the target and reference images, ensuring that elements in the target image reference only the related regions in the reference image. It should be noted that this is an optional strategy, employed only when the re-generated targets significantly exceed the initial range.

\subsubsection{Masked Mutual Self-Attention}
To alleviate the confusion between multiple subjects, we propose that only the appearances between the same roles can be referenced, \ie, multiple subjects in the target image can only refer to the same corresponding subject in other reference images. 
Given $N$ subject portraits (reference image), we aim to generate one corresponding scene image (target image). By constructing the subject mask, the formalization of our self-attention layer is as follows,
\begin{align}
    K^{+} &= [K_1 \oplus K_2 \oplus \ldots \oplus K_{N} \oplus K_\text{\scriptsize \tiny TGT}],  \\
    V^{+} &= [V_1 \oplus V_2 \oplus \ldots \oplus V_{N} \oplus V_\text{\scriptsize \tiny TGT}],  \\
    M^{+} &= [M_1 \oplus M_2 \oplus \ldots \oplus M_{N} \oplus \mathds{1}], \\
    A^{+} &= \textit{softmax}\left({Q_\text{\scriptsize \tiny TGT} K^{+}}/{\sqrt{d_k}} + \log M^{+} \right),  \\
    O_\text{\scriptsize \tiny TGT} &= \textit{conv}_{\scriptsize \tiny out}(A^{+} \cdot V^{+}),
\end{align}

where $M_{i}$ is the subject mask for \textit{i}-th reference images, and $\oplus$ indicates the concatenation operation. We assume the last one to be the target image, denoted with the subscript ‘\textit{TGT}’. The standard attention masking technique is adopted, which nullifies softmax's logits by assigning their scores to $-\infty$ based on the mask, followed by previous works~\cite{cao2023masactrl, tewel2024_ConsiStory}. It should be noted that, unlike ConsiStory~\cite{tewel2024_ConsiStory} and StoryDiffusion~\cite{storydiffusion}, which allow all areas of the target image to reference the subject in the reference image, our method only permits referencing information from the same subject.

\subsubsection{Masked Mutual Cross-Attention} 
As mentioned above, the rich information about the subject is not only contained in the reference image but also in the reference text. To fully utilize this information, we’ve implemented a Masked Mutual Cross-Attention (MMCA) mechanism. Its core idea is to allow the subject in the target image to query their reference text embedding and obtain rich, detailed attributes. We replace $K$ and $V$ with those of the corresponding reference. Meanwhile, a subject mask ensures that only the subject area of the target image will query the corresponding reference text embedding. To enhance the stability, we adopt fusing multiple frames of information before adding the residuals. Therefore, our cross-attention can be formalized as follows,
\begin{align}
    A_\text{\scriptsize \tiny TGT}^{i} &= \textit{softmax}\left({Q_\text{\scriptsize \tiny TGT} K_i}/{\sqrt{d_k}} + \log (m_\text{\scriptsize \tiny TGT}^{i} \times \mathds{1}) \right),  \\
    O_\text{\scriptsize \tiny TGT}^{i} &= \textit{conv}_{\scriptsize \tiny out} \left( A_\text{\scriptsize \tiny TGT}^{i} \cdot V_i \right).
\end{align}
Subsequently, all the $O_\text{\scriptsize \tiny TGT}^{i}$ are accumulated in a mask-weighted manner to reduce the confusion of different subjects. For the overlapping area of the mask, we take the average value. The calculation can be formulated as follows,

\begin{align}
    O_\text{\scriptsize \tiny TGT} &= \lambda \frac{m_{u}}{m_{s}} \sum_{i=0}^{N} O_\text{\scriptsize \tiny TGT}^{i}  + O^\text{\scriptsize \tiny vanilla}_\text{\scriptsize \tiny TGT} * (1-m_{u}) * (1-\lambda),
\end{align}
where the $m_{s}$ and $m_{u}$ respectively represent the sum and intersection of $m_i$ ($i \in [1,N]$) and $\lambda$ is the weight of text feature injection.
In practice, we add a small number ($10^{-8}$) to $m_{s}$ to prevent division by zero errors. The symbol $O^\text{\scriptsize \tiny vanilla}_\text{\scriptsize \tiny TGT}$ indicates the output of the target image from the vanilla forward process, which contains much background information of the target image.

This mechanism ensures that each subject in the scene image references only its corresponding text, thus obtaining a wealth of attributes. Such a strategy is vital for extracting text information from the reference anchor, which in turn aids in the generation of a vivid and precise story visualization.

\begin{figure*}[t]
\setlength{\abovecaptionskip}{2pt} 
\setlength{\belowcaptionskip}{-0.2cm} 
\centering
\includegraphics[width=1.0\linewidth]{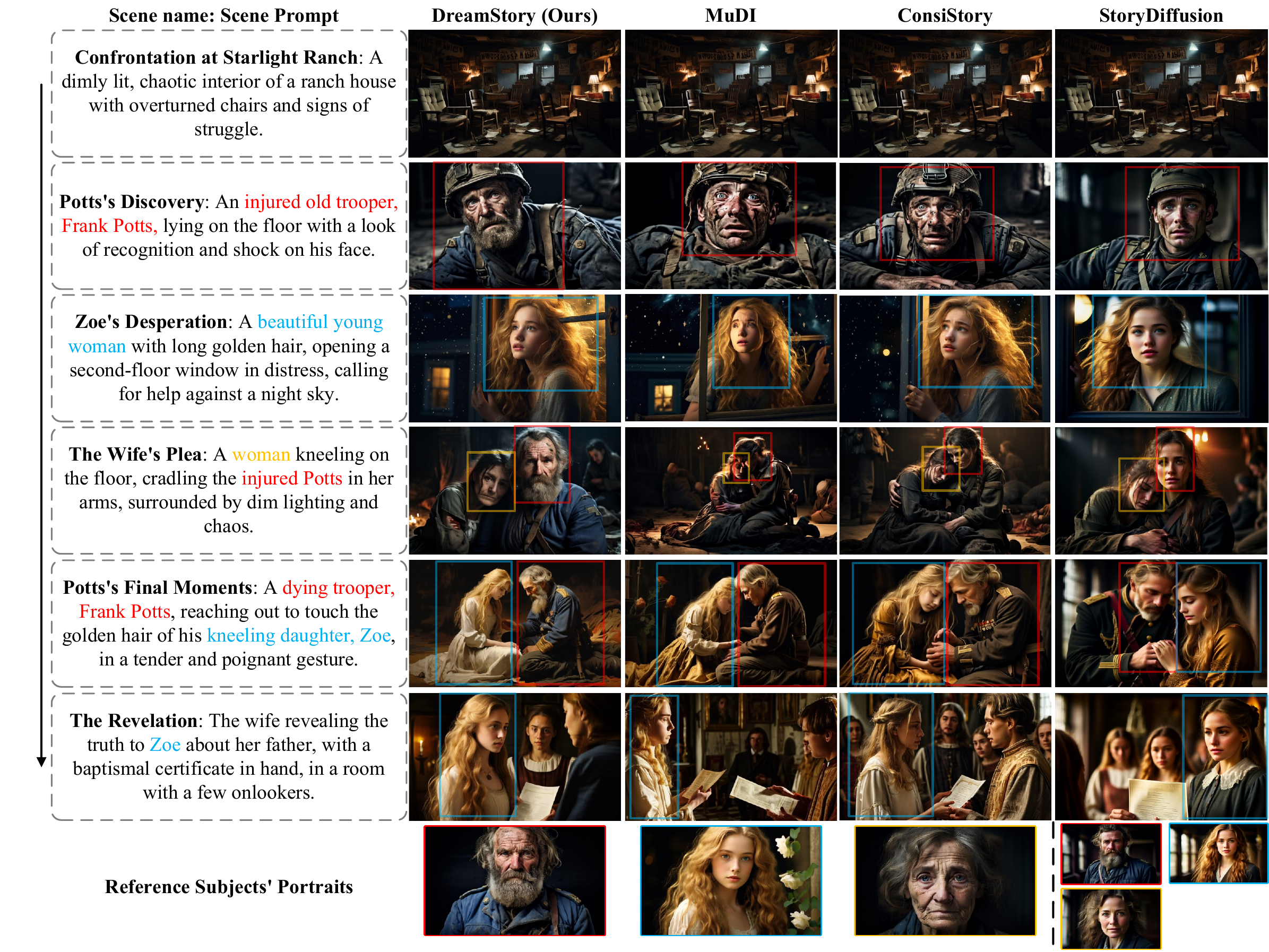}
\caption{Qualitative comparisons of our DreamStory with SOTA approaches on the FSS real story benchmark. 
        Ours, MuDI, and ConsiStory utilize the subject image on the bottom-left as the reference image. In contrast, StoryDiffusion references the subject image on the bottom-right.
        Different subjects are indicated with different colors. Please visit the project \href{https://dream-xyz.github.io/dreamstory}{homepage} to watch the video.
        }
\label{fig:storyline_FFS}
\vspace{-0.3cm}
\end{figure*}

\begin{figure*}[t]
\setlength{\abovecaptionskip}{2pt} 
\setlength{\belowcaptionskip}{-0.2cm} 
\centering
\includegraphics[width=1.0\linewidth]{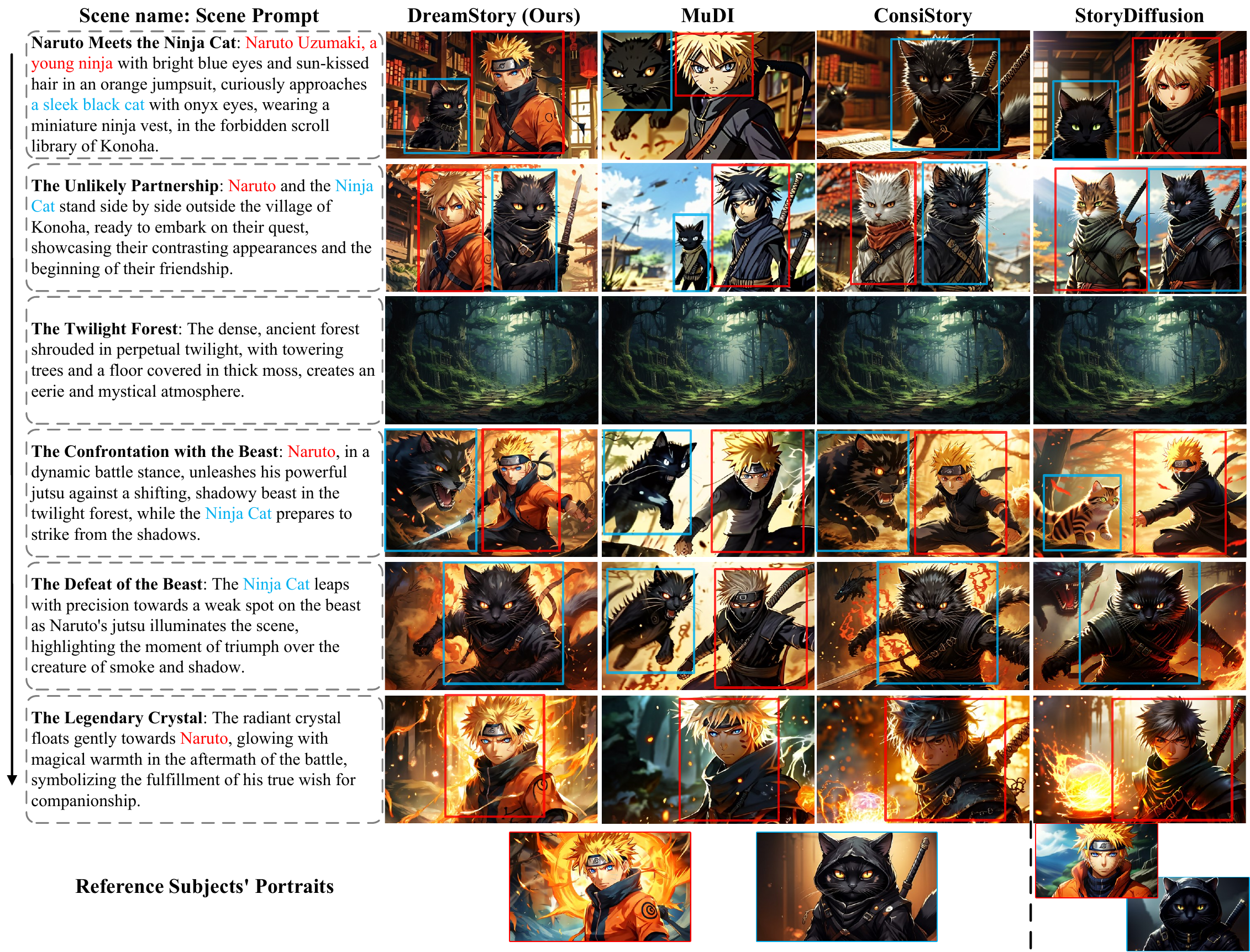}
\caption{Qualitative comparisons of our DreamStory with SOTA approaches on the ChatGPT generated story benchmark. 
        Ours, MuDI, and ConsiStory utilize the subject image on the bottom-left as the reference image. In contrast, StoryDiffusion references the subject image on the bottom-right.
        Different subjects are indicated with different colors. Please visit the project \href{https://dream-xyz.github.io/dreamstory}{homepage} to watch the video.
        }
\label{fig:storyline_2IP}   
\vspace{-0.3cm}
\end{figure*}

\section{Experiments}\label{sec:exp}

This section will introduce the evaluation benchmark and metrics, implementation details, and comprehensive experimental results. 
In Sec.~\ref{subsec:exp_benchmark}, we will introduce the constructed benchmark, which includes 100 stories and 400 synthetic cases. Sec.~\ref{subsec:exp_metrics} will introduce the objective and subjective evaluation metrics.
We present the specific implementation details in Sec.~\ref{subsec:exp_implementation_details}, including our LLM, diffusion backbone, and MSD module. In Sec.~\ref{subsec:exp_compare_SOTA} and Sec.~\ref{subsec:exp_ablation}, we will respectively present the comparative results with the current state-of-the-art (SOTA) methods and conduct an ablation study. Sec.~\ref{subsec:exp_limitation} briefly discusses the limitations of our method.

\subsection{Evaluation Benchmark} \label{subsec:exp_benchmark}

To our knowledge, few datasets can validate the proposed \textbf{DreamStory}’s performance in open-domain story visualization. To address this issue, we constructed a benchmark \textbf{DS-500}, including 100 real stories and 400 synthetic cases.
The benchmark of 100 real stories assesses our framework's holistic performance. The additional 400 synthetic samples, divided into four groups of 100 samples, each with 0, 1, 2, and 3 subjects, are utilized to evaluate the precision of the LLM in annotating subjects present in the scene and the efficacy of multi-subject consistent generation.

\subsubsection{The 100 Stories Benchmark} 
To validate the effectiveness of our overall framework, we first constructed a dataset. This dataset consists of 50 real, copyright-free English stories randomly downloaded from 
\textit{free-short-stories}~\footnote{\href{https://theshortstory.co.uk/resources/free-short-stories/}{Free Short Stories}}, and 50 short stories generated by ChatGPT. These data effectively simulate the distribution of real stories, thereby providing a robust validation of the performance of our \textbf{DreamStory} in open-domain story visualization.

\subsubsection{The 400 Synthetic Benchmark} 
Firstly, we instruct GPT to generate a variety of non-repetitive subjects, each accompanied by detailed portrait prompts. We then employ GPT to annotate these subjects with type attributes (\eg, girl, man, dog), which are applicable for DINO detection. Subsequently, a subset of subjects is randomly selected, and GPT is tasked to generate scene prompts that exclusively include the chosen subjects. To prevent performance degradation of the diffusion model due to overly lengthy output text, we limit GPT’s output to approximately 40 words (roughly 50 tokens). These scene prompts, along with their associated subject prompts and type attributes, constitute the 400 synthetic benchmarks. Finally, these datasets will be manually checked and filtered to ensure accuracy.

\begin{table*}[t]
\setlength{\abovecaptionskip}{1pt} 
\setlength{\belowcaptionskip}{0.4cm}
    \centering
    \small \renewcommand{\arraystretch}{1.25}
    \caption{Quantitative results of different backbones for our DreamStory on the DS-500 benchmark. {\color{red}{\textbf{Red}}} indicate the best performance.}
    \label{tab:results_ablation_backbone}\

\begin{tabular}{p{3.2cm}| c c c c | c c c c}
\bottomrule[1.5pt]
\multicolumn{1}{l|}{\multirow{2}{*}{}}                               & \multicolumn{4}{c|}{\textbf{2-Subject}} & \multicolumn{4}{c}{\textbf{3-Subject}} \\
\multicolumn{1}{l|}{}                                               & AES$\uparrow$ & CLIP-T$\uparrow$  & DS$\uparrow$  &  D\&C-DS$\uparrow$   & AES$\uparrow$ & CLIP-T$\uparrow$  & DS$\uparrow$  &  D\&C-DS$\uparrow$   \\ \hline
SDXL~\cite{podell2023sdxl}         & 6.52    & {\color{red}{\textbf{0.3819}}}   & 0.5045   &  0.3018    & 6.59   & {\color{red}{\textbf{0.3900}}} & 0.4618   & 0.1241 \\
SDXL~\cite{podell2023sdxl} + \textbf{Ours}  & 6.62 & 0.3747& 0.6048 & 0.3848 & 6.69 & 0.3832 & 0.5228 & 0.1778 \\ \hline \hline
Playground~\cite{li2024playground}         & 6.67    & 0.3818   & 0.5796   & 0.3996  & 6.77   & 0.3841   & 0.5194   & 0.1938   \\
Playground~\cite{li2024playground} + \textbf{Ours} & {\color{red}{\textbf{6.72}}}& 0.3779   &{\color{red}{\textbf{0.6714}}}& {\color{red}{\textbf{0.5444}}} &{\color{red}{\textbf{6.81}}}& 0.3791   &{\color{red}{\textbf{0.5965}}}& {\color{red}{\textbf{0.2335}}}  \\ \toprule[1.5pt]
\end{tabular}
\vspace{-10pt}
\end{table*}

\begin{figure*}[t]
\centering
\setlength{\abovecaptionskip}{2pt} 
\setlength{\belowcaptionskip}{-0.2cm} 

\includegraphics[width=1.0\linewidth]{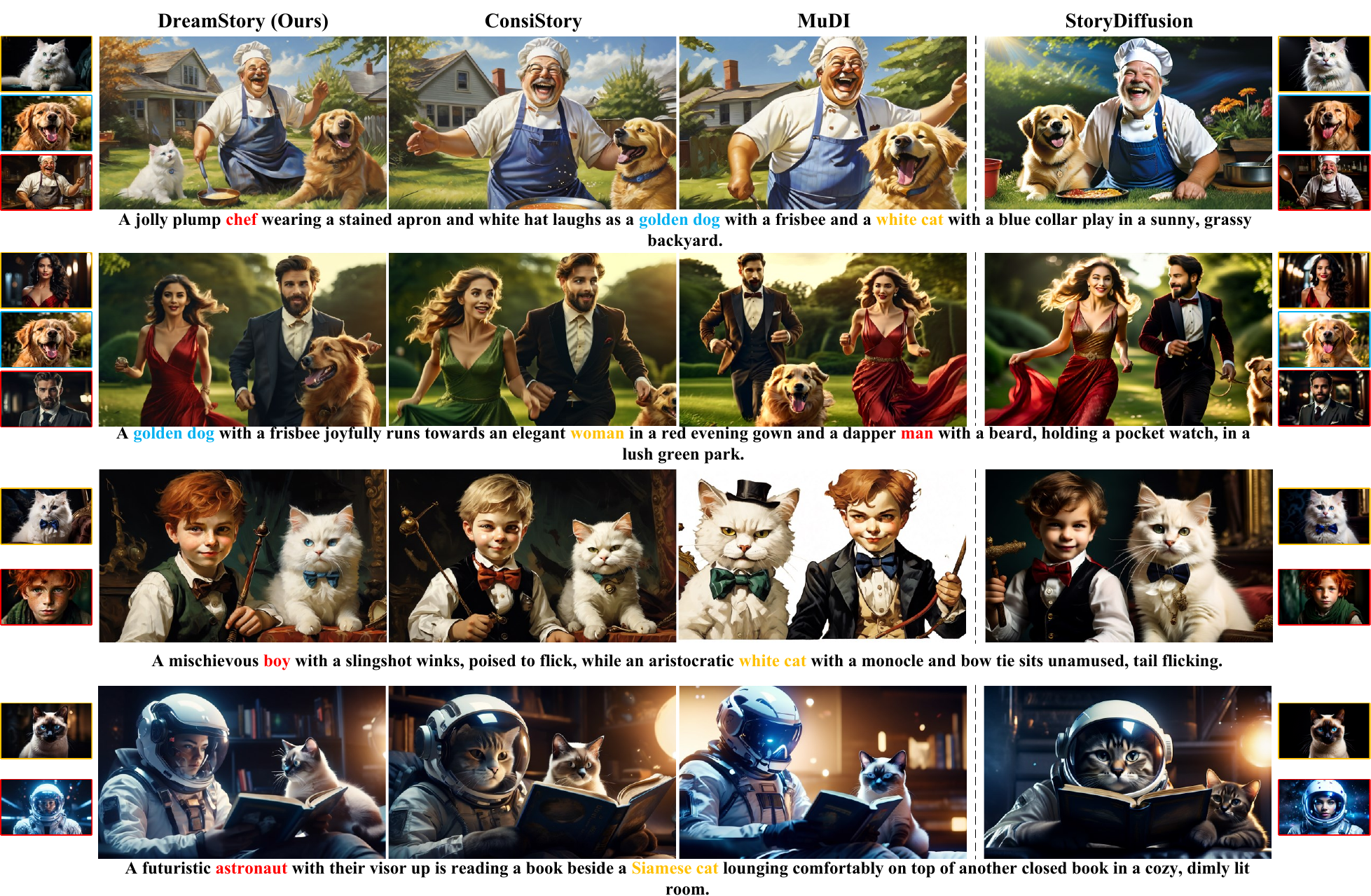}
\centering
\caption{Qualitative comparisons of our DreamStory with SOTA approaches on the synthetic benchmark. 
        Ours, MuDI, and ConsiStory utilize the subject image on the left as the reference image. In contrast, StoryDiffusion references the subject image on the right.
        Different subjects are indicated with different colors.
        Our method better maintains consistency across multiple subjects, such as the cat in the first row, the color of the man's suit and the woman's hair in the second row, the hair color of the boy in the third row, and the head of the astronaut in the fourth row.}
\label{fig:benchmark_vis}
\vspace{-0.3cm}
\end{figure*}

\begin{table*}[t] \small
    \setlength{\abovecaptionskip}{1pt} 
    \setlength{\belowcaptionskip}{0.4cm}
    \centering
    \small \renewcommand{\arraystretch}{1.25}
    \caption{Quantitative comparison on benchmark. {\color{red}{\textbf{Red}}} and {\color{blue}{\uline{blue}}} indicate the best and the second-best performance.}
    \label{tab:results_benchmark}

\begin{tabular}{p{3cm}| c c c c | c c c c}
\bottomrule[1.5pt]
\multicolumn{1}{l|}{\multirow{2}{*}{}}                                                 & \multicolumn{4}{c|}{\textbf{2-Subject}} & \multicolumn{4}{c}{\textbf{3-Subject}} \\
\multicolumn{1}{l|}{}                                                                  & AES$\uparrow$ & CLIP-T$\uparrow$  & DS$\uparrow$  &  D\&C-DS$\uparrow$   & AES$\uparrow$ & CLIP-T$\uparrow$  & DS$\uparrow$  &  D\&C-DS$\uparrow$   \\ \hline
MuDI~\cite{MUDI}                                & 6.47    & 0.3652   & {\color{blue}{\uline{0.6578}}}     &  {\color{blue}{\uline{0.4410}}}    &   6.54   &  0.3664   &  {\color{blue}{\uline{0.5924}}}  &  0.1988  \\
ConsiStory~\cite{tewel2024_ConsiStory}  & {\color{blue}{\uline{6.62}}}    & {\color{blue}{\uline{0.3757}}}   & 0.5988   & 0.4251  & {\color{blue}{\uline{6.73}}}   & {\color{blue}{\uline{0.3770}}}   & 0.5564   & 0.2038  \\
StoryDiffusion~\cite{storydiffusion}    & 6.56    & 0.3702   &      0.6258   &      0.4364  & 6.57   & 0.3707   & 0.5723   & {\color{blue}{\uline{0.2095}}}     \\
\textbf{DreamStory} (\textbf{Ours})              & {\color{red}{\textbf{6.72}}}    & {\color{red}{\textbf{0.3779}}}   & {\color{red}{\textbf{0.6714}}}   &   {\color{red}{\textbf{0.5444}}} & {\color{red}{\textbf{6.81}}}   & {\color{red}{\textbf{0.3791}}}   & {\color{red}{\textbf{0.5965}}}   &  {\color{red}{\textbf{0.2335}}} \\ \toprule[1.5pt]
\end{tabular}
\vspace{-10pt}
\end{table*}

\subsection{Evaluation Metrics}  \label{subsec:exp_metrics}
In story visualization, aesthetics and image-text alignment are commonly employed metrics. In addition, the consistency of subjects across multiple frames is another crucial metric, which is one of the main problems this paper aims to address. Therefore, we evaluate generated results using three criteria: 1) \textbf{aesthetics}, 2) \textbf{consistency between scene image and text}, and 3) \textbf{subject consistency between scene and reference image}. To ensure accuracy and reliability, each criterion is evaluated objectively and subjectively.

\subsubsection{Objective Evaluation} 
Followed by previous works~\cite{podell2023sdxl, li2024playground}, we utilize an aesthetic predictor~\footnote{\href{https://github.com/christophschuhmann/improved-aesthetic-predictor}{improved-aesthetic-predictor}} to determine aesthetic scores. The CLIP~\footnote{\href{https://huggingface.co/openai/clip-vit-base-patch16}{clip-vit-base-patch16}} score is adopted to evaluate the similarity between the scene text and scene image, denoted as CLIP-T. 
To better assess subject consistency, we employ DreamSim~\cite{fu2024dreamsim} to evaluate the similarity between two subject images. The GroundingDINO~\cite{GroundingDINO} is first applied to detect the bounding box of the target subject based on its category, \eg, man or dog. For each subject, we use the image cropped from the highest-probability bounding box to compute DreamSim similarity. This original DreamSim score is denoted as \textbf{DS}. 
Furthermore, generating multiple subjects makes it feasible to create composite subjects that blend elements from multiple others. This can lead to a single composite subject scoring high in \textbf{DS} with multiple subjects. Therefore, we adopted the \textbf{D\&C-DS}~\cite{MUDI} metric to evaluate the consistency across multiple subjects, which has been validated to align with human preference. We also evaluate the accuracy of LLM annotation in 400 synthetic benchmarks.

\subsubsection{Subjective Evaluation} 
Due to the bias of existing metrics, a user study is conducted to assess subjective results. Given the variability of individual ratings and the broad spectrum of scores across different evaluators, we employed a pairwise comparison in our user study. For each evaluation, two sets of images were randomly displayed, each generated by a different method and accompanied by its respective text. Participants were asked to judge each metric by selecting one of three options: Image A is superior, Image B is superior, or both are comparable.
We engaged 20 independent evaluators for the assessment. Each evaluator conducted 100 reviews per benchmark, culminating in 2000 votes in total. The final results were compiled and are presented as percentages.

\begin{figure*}[t]
\centering
\setlength{\abovecaptionskip}{2pt}
\setlength{\belowcaptionskip}{-0.4cm}

\includegraphics[width=1.0\linewidth]{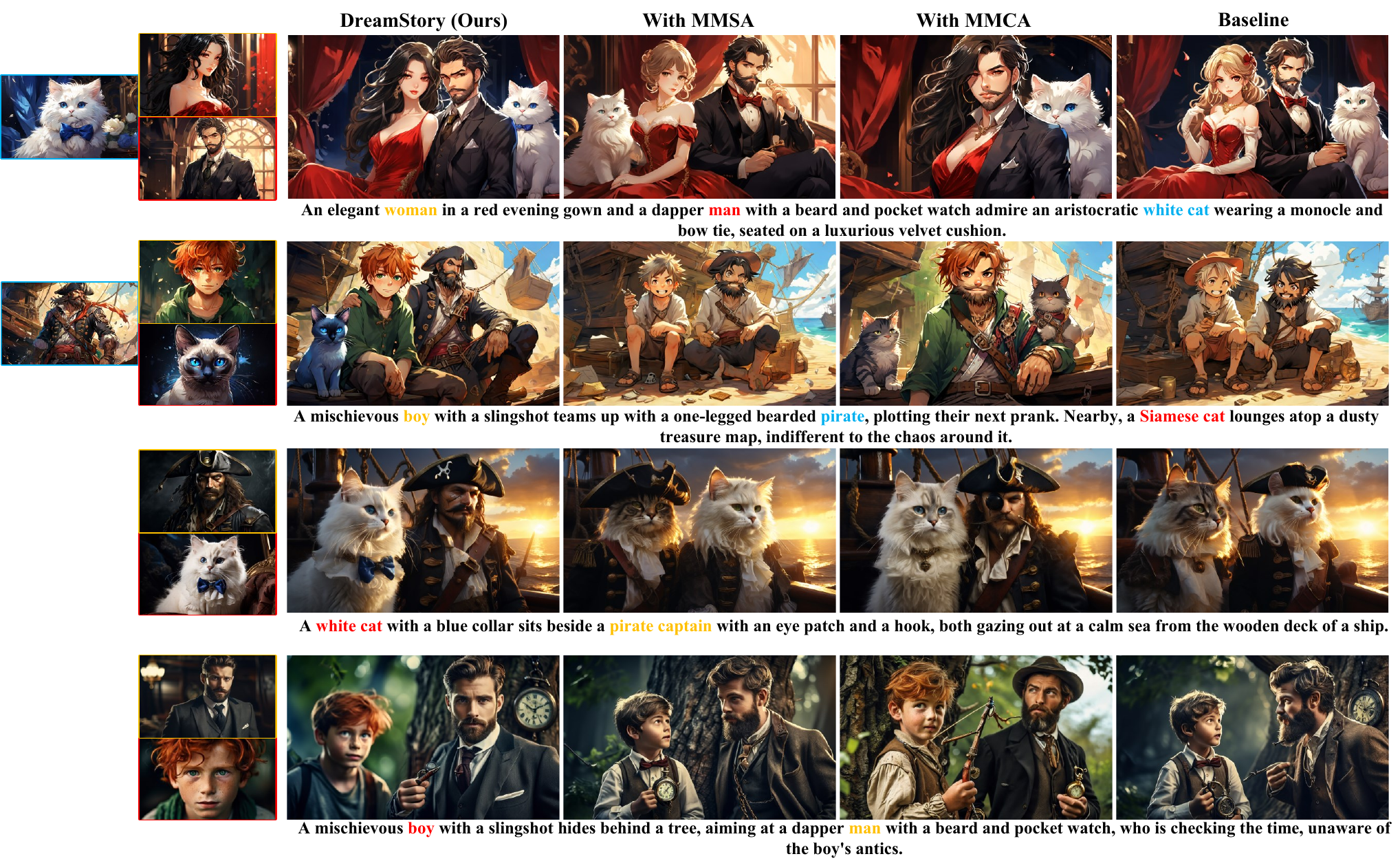}
\centering
\caption{Ablation studies of different generations results. 
        All settings except the baseline utilize the subject image on the left as the reference image.
        Different subjects are indicated with different colors.
        Our method better maintains consistency across multiple subjects, such as the woman and cat in the first row, 
        the pirate and Siamese cat in the second row, 
        the pirate and cat in the third row, 
        and the boy in the last row.}
\label{fig:benchmark_ablation}
\end{figure*}

\begin{table*}[t]
\setlength{\abovecaptionskip}{0.2cm} 
\setlength{\belowcaptionskip}{0.4cm}
    \centering
    \small \renewcommand{\arraystretch}{1.25}
    \caption{Quantitative results of ablation study on the benchmark. {\color{red}{\textbf{Red}}} and {\color{blue}{\uline{blue}}} indicate the best and the second-best performance.}
    \vspace{-6pt}
    \label{tab:results_ablation_benchmark}\

\begin{tabular}{p{3.8cm}| c c c c | c c c c}
\bottomrule[1.5pt]
\multicolumn{1}{l|}{\multirow{2}{*}{}}                              & \multicolumn{4}{c|}{\textbf{2-Subject}} & \multicolumn{4}{c}{\textbf{3-Subject}} \\
\multicolumn{1}{l|}{}                                               & AES$\uparrow$ & CLIP-T$\uparrow$  & DS$\uparrow$  &  D\&C-DS$\uparrow$   & AES$\uparrow$ & CLIP-T$\uparrow$  & DS$\uparrow$  &  D\&C-DS$\uparrow$   \\ \hline

Baseline         & 6.67    & {\color{red}{\textbf{0.3818}}}   & 0.5796   & 0.3996  & 6.77   & {\color{blue}{\uline{0.3841}}}   & 0.5194   & 0.1938   \\
w/ MMCA          & 6.68   & 0.3791   &{\color{blue}{\uline{0.6673}}}   &  {\color{blue}{\uline{0.5301}}}    &{\color{blue}{\uline{6.80}}}& 0.3772   &{\color{blue}{\uline{0.5888}}} & {\color{blue}{\uline{0.2186}}} \\
w/ MMSA            & {\color{blue}{\uline{6.69}}}    &{\color{blue}{\uline{0.3800}}}   & 0.5922   & 0.4233  & 6.76   & {\color{red}{\textbf{0.3852}}}   & 0.5293   & 0.2098 \\
w/ MMSA+MMCA (\textbf{Ours}) & {\color{red}{\textbf{6.72}}}& 0.3779   &{\color{red}{\textbf{0.6714}}}& {\color{red}{\textbf{0.5444}}} &{\color{red}{\textbf{6.81}}}& 0.3791   &{\color{red}{\textbf{0.5965}}}& {\color{red}{\textbf{0.2335}}}  \\ \toprule[1.5pt]
\end{tabular}
\vspace{-10pt}
\end{table*}

\begin{figure*}[t]
\setlength{\abovecaptionskip}{2pt} 
\setlength{\belowcaptionskip}{-0.2cm} 
\centering
\includegraphics[width=1.0\linewidth]{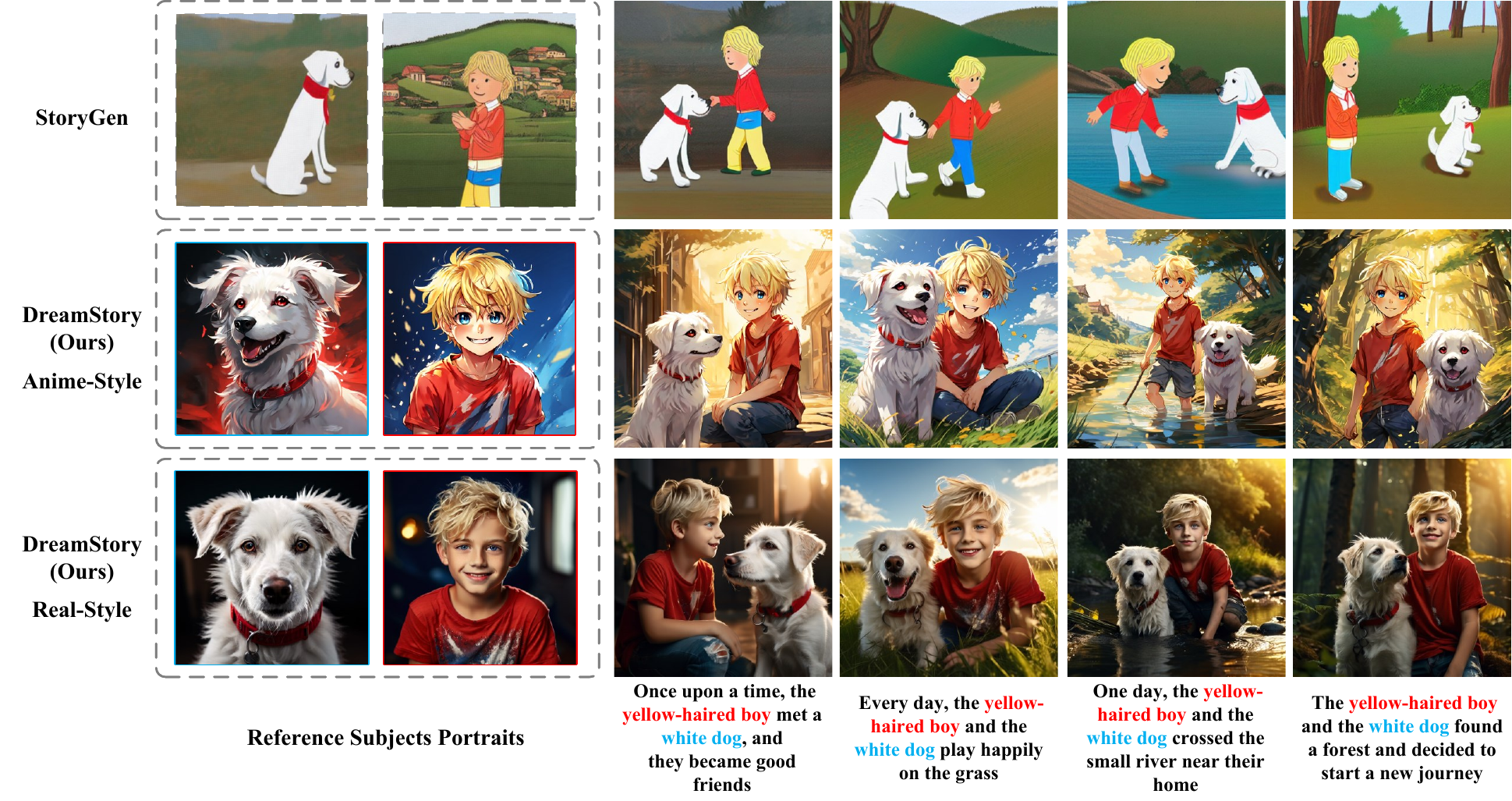}
\caption{Qualitative comparisons of our DreamStory with StoryGen on their benchmark. All approaches utilize the
subject image on the left as the reference image. Different subjects are indicated with different colors. 
The narrative text of the story is presented below and serves as the input for our DreamStory.
Please visit the project \href{https://dream-xyz.github.io/dreamstory}{homepage} to watch the video.
}
\label{fig:storygen}
\vspace{-10pt}
\end{figure*}

\begin{figure}[t]
\setlength{\abovecaptionskip}{2pt} 
\setlength{\belowcaptionskip}{-0.2cm} 
\centering
\includegraphics[width=0.98\linewidth]{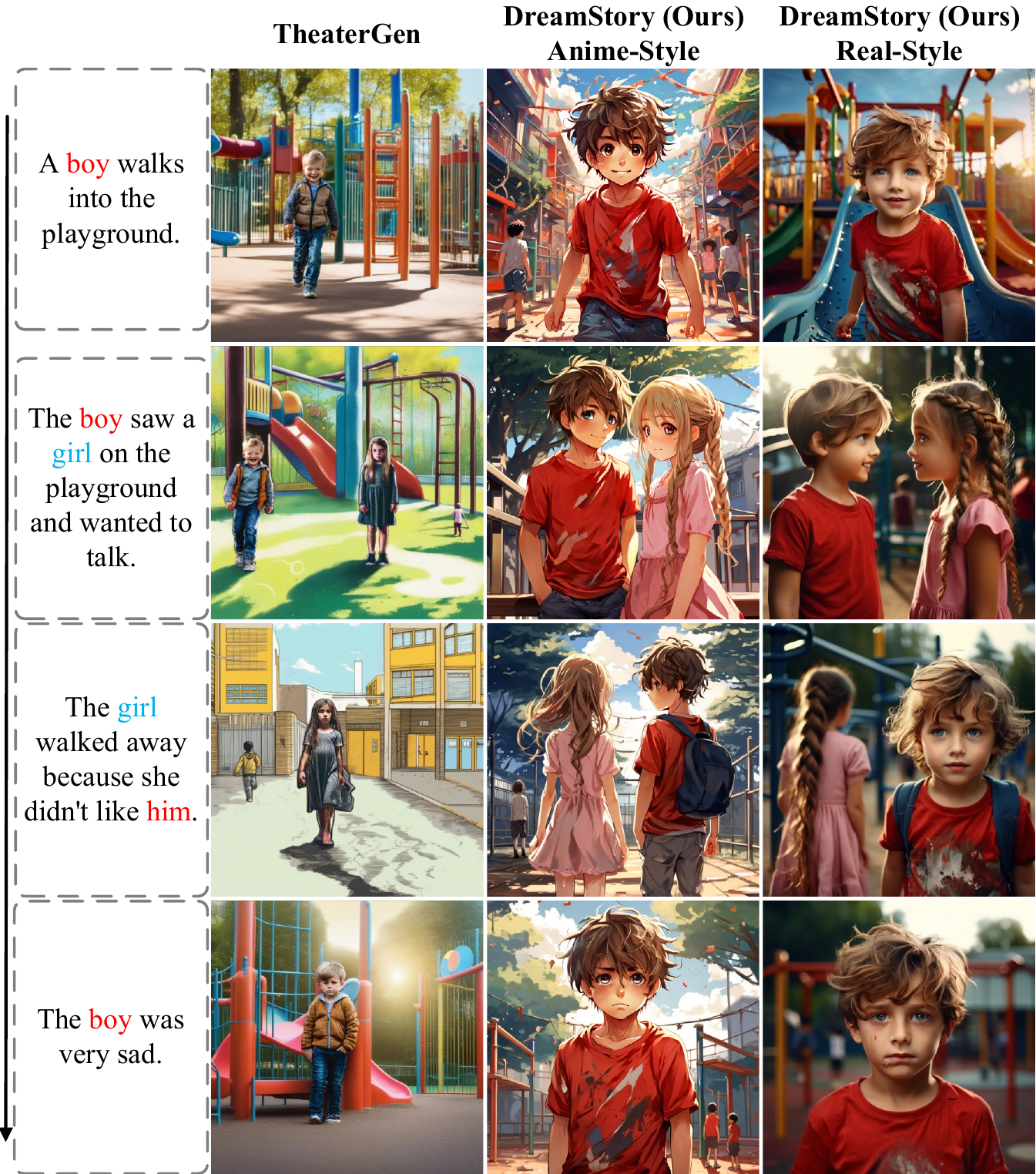}
\caption{Qualitative comparisons of our DreamStory with TheaterGen on their benchmark. 
    Different subjects are indicated with different colors. 
    The narrative text of the story is presented on the left and serves as the input for our DreamStory.
    Please visit the project \href{https://dream-xyz.github.io/dreamstory}{homepage} to watch the video.
    }
\label{fig:TheaterGen}
\end{figure}

\begin{figure}[t]
\setlength{\abovecaptionskip}{2pt} 
\setlength{\belowcaptionskip}{-0.2cm} 
\centering
\includegraphics[width=0.98\linewidth]{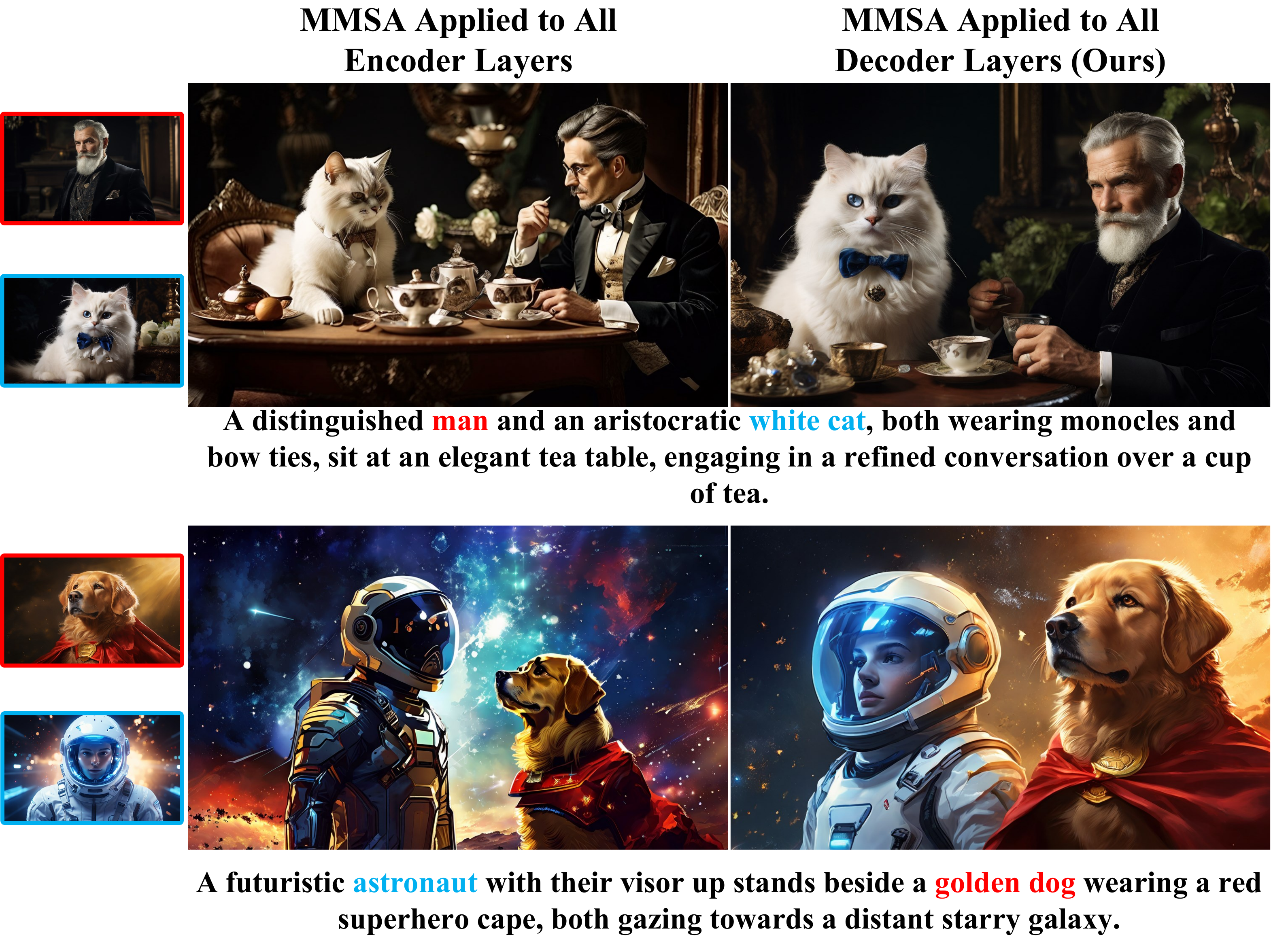}
\caption{Qualitative comparison of applying MMSA to U-Net encoder versus decoder layers. Different subjects are indicated with different colors. The results demonstrate that only applying MMSA to decoder layers preserves appearance consistency.}
\label{fig:ABL_MMSA}
\vspace{-10pt}
\end{figure}

\begin{figure*}[t]
\centering
\setlength{\abovecaptionskip}{2pt} 
\setlength{\belowcaptionskip}{-0.2cm} 
\includegraphics[width=0.98\linewidth]{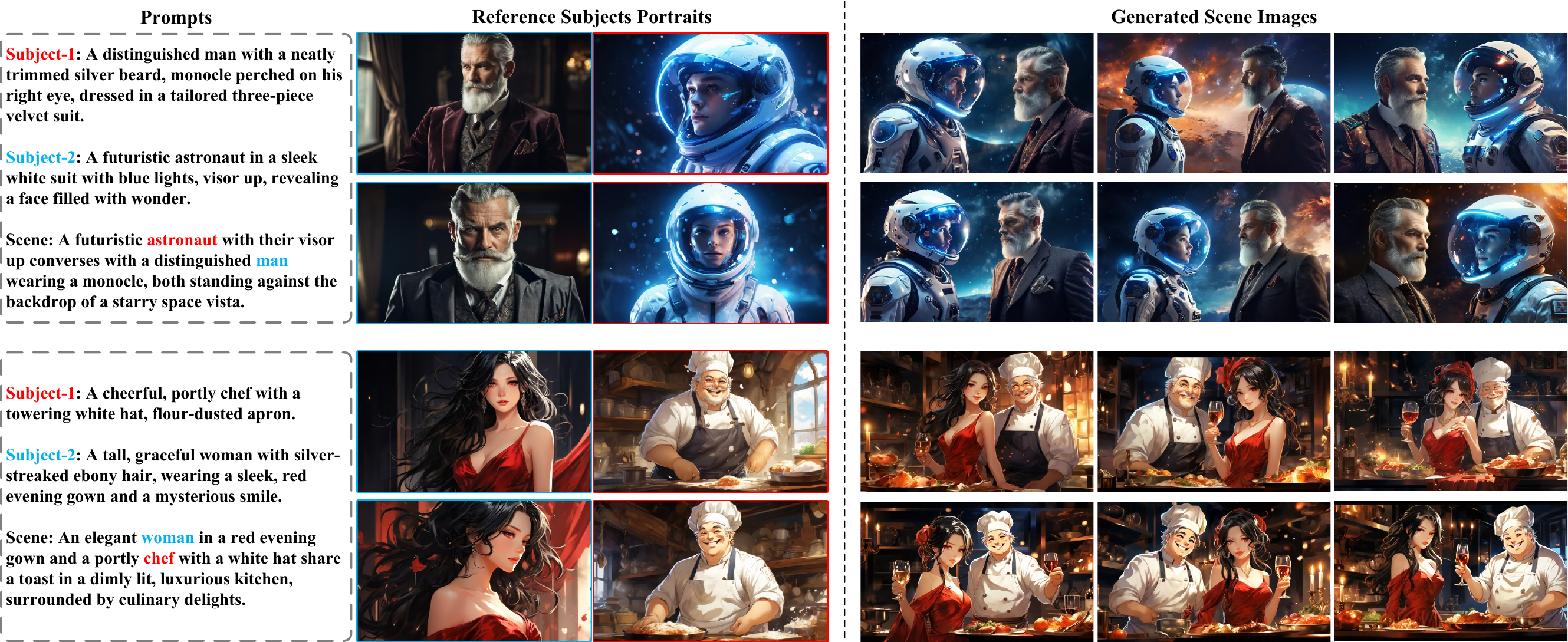}
\centering
    \caption{Qualitative results of our DreamStory with a different random seed. 
    The first two rows display the same set of subjects in real style with different seeds; the last two rows present another set in anime style. Each row contains two subject images on the left and three scene images on the right, which are generated based on the left subjects.
    Different subjects are indicated with different colors. 
    These results demonstrate the superior consistency and diversity of our DreamStory across various styles.
    }
\label{fig:diversity}
\end{figure*}

\begin{figure*}[t]
\setlength{\abovecaptionskip}{2pt} 
\setlength{\belowcaptionskip}{-0.2cm} 
\centering
\includegraphics[width=0.98\linewidth]{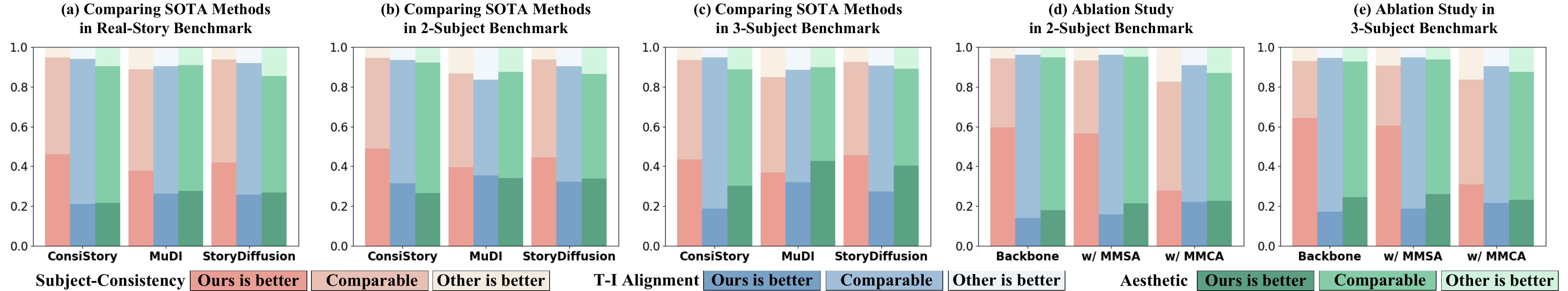}
\caption{User Study on DS-500 benchmark. 
        Dominant preferences to our full model are presented, compared with other competitive baselines (a, b, c) and ablation models (d, e).
        T-I Alignment means text-image relevance.}
\label{fig:user_study}
\end{figure*}

\subsection{Implementation Details} \label{subsec:exp_implementation_details}

\textbf{LLMs as Story Director.} We utilize ChatGPT4~\cite{openai2023GPT4}, currently the most advanced large-scale language model, as our \textit{story director} due to its powerful interactive and long context capabilities. Interaction with the LLM is conducted via their API~\footnote{\href{https://pypi.org/project/openai/}{openai:gpt-4-turbo-preview}}.

\textbf{Diffusion Model Backbone.} 
We first conducted an ablation study on two popular T2I backbones, Playground~\footnote{\href{https://huggingface.co/playgroundai/playground-v2.5-1024px-aesthetic}{playground-v2.5-1024px-aesthetic}} and SDXL~\footnote{\href{https://huggingface.co/stabilityai/stable-diffusion-xl-base-1.0}{SDXL-base-1.0}}. The results are presented in Tab.~\ref{tab:results_ablation_backbone}.
We adopt Playground as the final T2I backbone due to its excellent performance in aesthetics and subject consistency. We utilize the default scheduler (EDMDPMSolverMultistepScheduler~\cite{lu2022dpmnew}) with 50 inference steps to ensure optimal performance during the inference phase. The guidance scale~\cite{ho2021classifier} is set to 7.0 in our experiments. The weight of text feature injection, $\lambda$, is fixed to 0.9 for a tradeoff between scene semantics and consistency of subjects. 
The evaluation of all our models focuses on generating visual content with dimensions of 1280 (width) by 768 (height).

\textbf{Attention Mechanisms in MSD.} Our MSD is applied across all diffusion steps to ensure multi-subject consistency. The Masked Mutual Self-Attention (MMSA) is applied to all decoder layers to maintain the appearance consistency. As shown in Fig.~\ref{fig:ABL_MMSA}, this strategy more effectively preserves the appearance consistency compared to applying MMSA to encoder layers. These observations align with prior works~\cite{cao2023masactrl,tewel2024_ConsiStory}, which highlight that the U-Net's decoder is specialized to handle fine-grained visual details (e.g., colors and textures), while the encoder manages structural information.
Inspired by previous work~\cite{controlnet, layoutWACV}, the Masked Mutual Cross-Attention (MMCA) is applied to all layers for better cross-attention fusion. The dropout~\cite{tewel2024_ConsiStory} strategy with a dropout rate of 0.5 is adopted to enhance layout diversity. Furthermore, we adopt the open-vocabulary segmentation model, GroundingSAM~\cite{GroundedSAM}, to generate precise masks for the subjects. This process begins with the detection of the subject using the open-vocabulary detection model, GroundingDINO~\cite{GroundingDINO}, followed by segmentation with the powerful SAM~\cite{SAM}. 

\subsection{Comparison with SOTA Methods} \label{subsec:exp_compare_SOTA}

To demonstrate the advantage of the proposed DreamStory, we compare our DreamStory with the state-of-the-art approaches. These methods fall into two main categories: (1) MuDI~\cite{MUDI}, fine-tuned using reference images; and (2) training-free methods, ConsiStory~\cite{tewel2024_ConsiStory} and StoryDiffusion~\cite{storydiffusion}. All the approaches are tested under the same setting for a fair comparison.

\subsubsection{Objective Comparison}
The overall results are presented in Tab.~\ref{tab:results_benchmark}. As can be seen from the table, our DreamStory outperforms other methods in terms of all metrics. Notably, the D\&C-DS metric of ours significantly surpassed other methods, exceeding MuDI, ConsiStory, and StoryDiffusion by margins of 0.1034(23.4\%), 0.1293(25.1\%), and 0.1080(24.7\%) respectively in the 2-Subject of DS-500 benchmark. This pattern is mirrored in the 3-Subject benchmark, reinforcing the effectiveness of our method in maintaining multi-subject consistency. Furthermore, our method exhibits a notable advantage on the DS metric, outperforming other training-free SOTA methods (ConsiStory and StoryDiffusion) on the 2-Subject benchmark by at least 0.05 (9.0\%). This lead, however, narrows to an approximate average of 0.02 (3.9\%) on the 3-Subject benchmark. This is attributed to the limitations of the diffusion model when generating three subjects, leading to a higher likelihood of subject fusion, which is also discussed in MuDI~\cite{MUDI}.

Regarding text similarity, all methods except MuDI~\cite{MUDI} yield comparable CLIP-T scores, and our method excels in all benchmarks. In contrast, MuDI has a significant CLIP-T decline (approximately 0.01) in all settings. This decline would worsen if the training continued, leading to problems such as lack of background or single subject dominating~\cite{MUDI}.
This suggests our method’s effectiveness in maintaining scene semantic consistency while preserving subject consistency.
In conclusion, all these results effectively prove that our DreamStory method performs superiorly in maintaining multi-subject consistency, enhancing aesthetics, and preserving scene semantic consistency. Notably, unlike MuDI, our training-free MSD eliminates the extra training costs and the potential for overfitting. This latter issue could notably degrade the aesthetic appeal of the generated scene images and their alignments with the corresponding text, which are crucial metrics in story visualization.

\begin{figure*}[t]
\centering
\setlength{\abovecaptionskip}{2pt} 
\setlength{\belowcaptionskip}{-0.2cm} 

\includegraphics[width=1.0\linewidth]{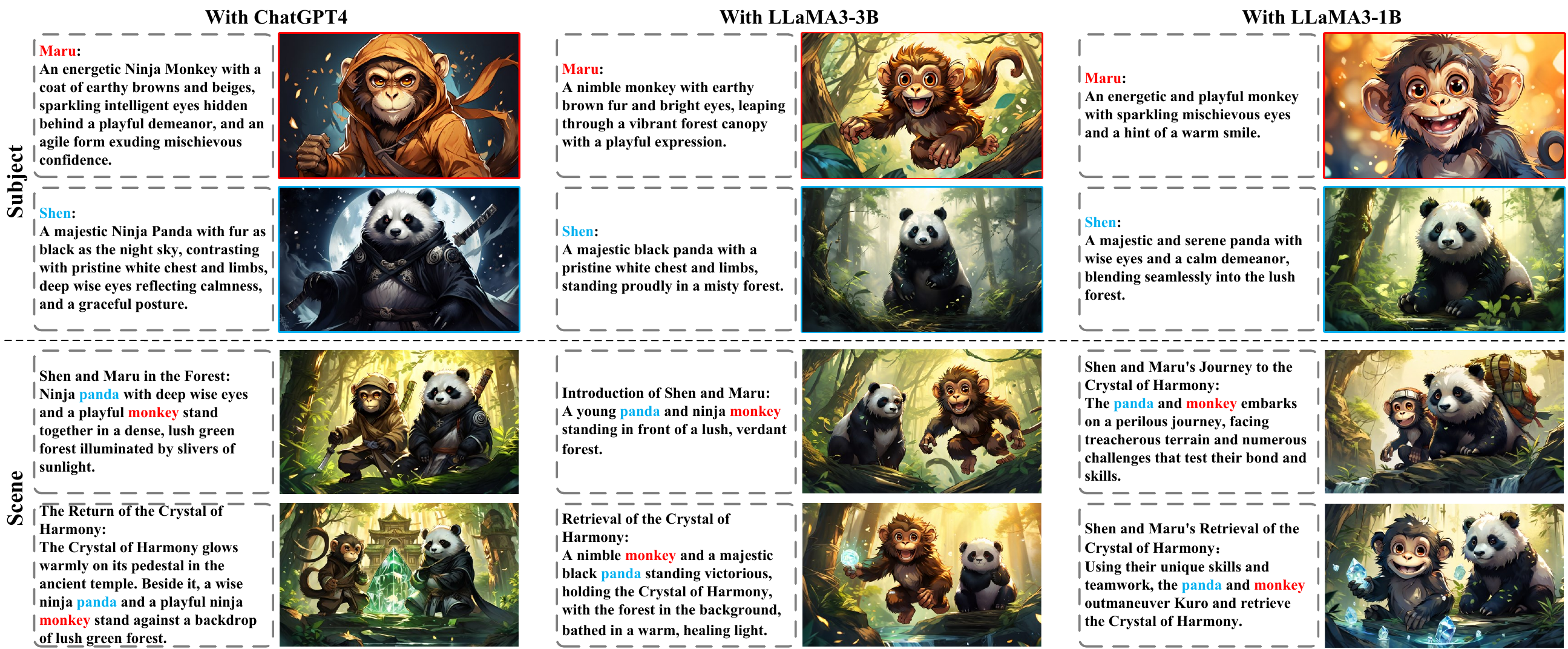}
\centering
    \caption{
    Qualitative results of our DreamStory with different LLMs (ChatGPT4, LLaMA3-3B, and LLaMA3-1B). The first two rows are the text and images of the subject, and the last two rows are those of the scenes. Different subjects are indicated with different colors.
    DreamStory generates visually appealing story scenes while maintaining consistency across multiple subjects, demonstrating the robustness of our framework.
    }
\label{fig:abl_llm}
\vspace{-10pt}
\end{figure*}

\begin{figure}[t]
\setlength{\abovecaptionskip}{2pt} 
\setlength{\belowcaptionskip}{-0.2cm}  
\centering
\includegraphics[width=1.0\linewidth]{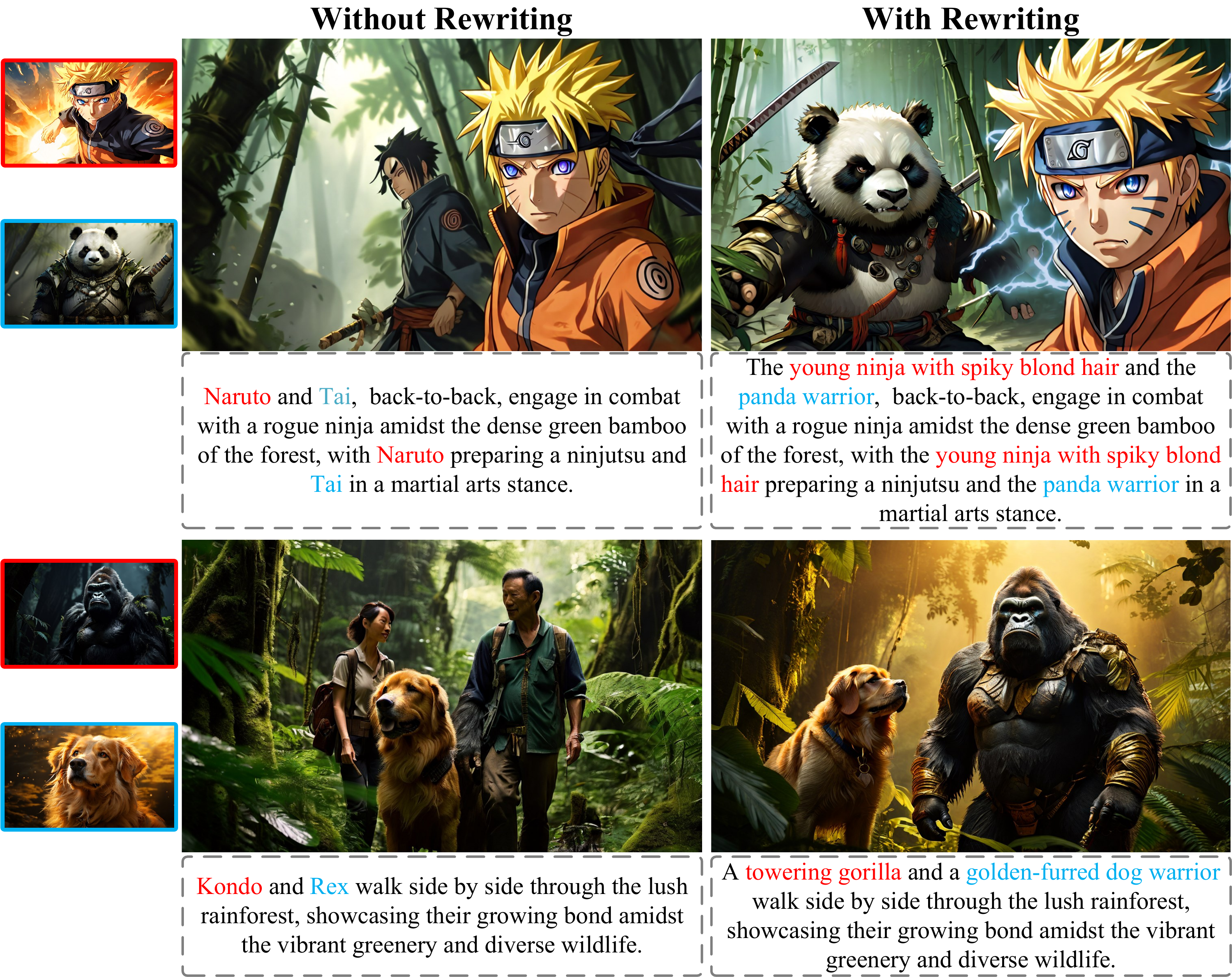}
\caption{Ablation study of LLM rewriting. Different subjects are indicated with different colors.
Without rewriting, the diffusion model may generate incorrect subjects, such as generating the panda Tai into a ninja man (first row), and turning the gorilla Kondo into a human (second row).}
\label{fig:LLM_rewriting}
\vspace{-10pt}
\end{figure}

\begin{figure}[t]
\setlength{\abovecaptionskip}{0.1cm} 
\setlength{\belowcaptionskip}{-0.2cm} 
\centering
\includegraphics[width=0.98\linewidth]{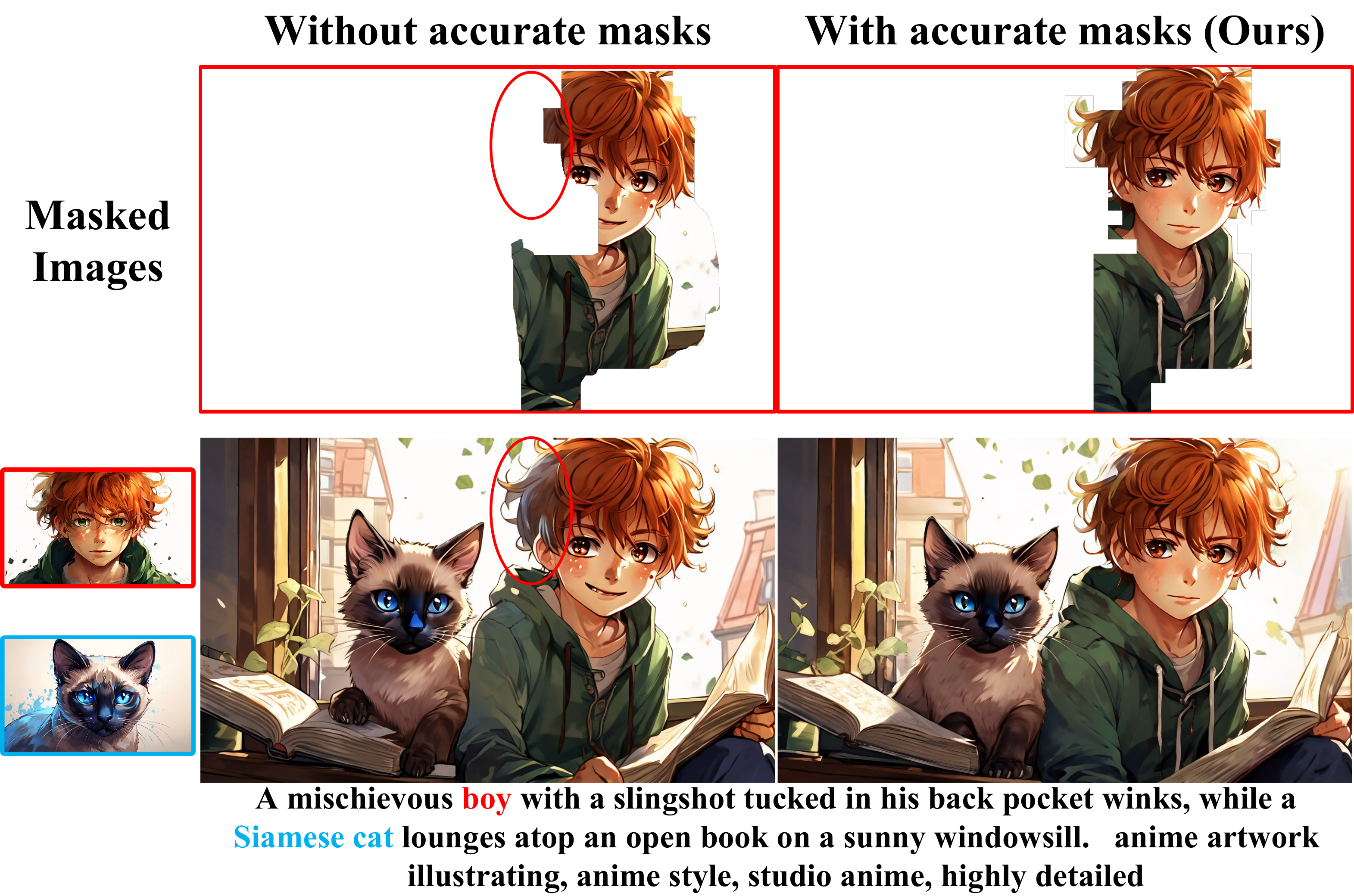}
\caption{Ablation study of accurate mask generation. Different subjects are indicated with different colors.
Without our accurate mask generation strategy, the initial mask (left) fails to fully cover the boy, causing inconsistent hair color (highlighted in red circle). }
\label{fig:ABL_mask_gen}
\vspace{-10pt}
\end{figure}

\subsubsection{Subjective Comparison} 
The overall subjective comparison results are presented in Fig.~\ref{fig:user_study}(a)(b)(c). It can be seen from Fig.~\ref{fig:user_study} that over 80\% evaluators believe that our DreamStory surpasses or is comparable to ConsiStory~\cite{tewel2024_ConsiStory}, MuDI~\cite{MUDI}, and StoryDiffusion~\cite{storydiffusion} in all benchmarks at all criteria. 
To demonstrate the performance of our overall framework, we present the visual results of two complete stories compared with other SOTA methods in Fig.~\ref{fig:storyline_FFS} and Fig.~\ref{fig:storyline_2IP}. From these tables, we can see that our framework is capable of effectively annotating the subjects in the scene that need to maintain consistency and generating images that maintain multi-subject consistency, such as the protagonist man and his daughter in Fig.~\ref{fig:storyline_FFS}, the Naruto and ninja cat in Fig.~\ref{fig:storyline_2IP}. To further illustrate the advantages of our approach in preserving multi-subject consistency, we also display more comparison visual results on the synthetic benchmark in Fig.~\ref{fig:benchmark_vis}. 

Furthermore, we apply our DreamStory to a case from StoryGen~\cite{storygen_liu2023intelligent}. We control the style of the generated results by adding style prompts, \eg, anime style, as shown in Fig.~\ref{fig:storygen}. As seen from Fig.~\ref{fig:storygen}, our method maintains better consistency while achieving a higher aesthetic appeal. Our training-free approach effectively leverages existing large, high-quality datasets (\eg, LAION-5B~\cite{schuhmann2022laion_5B}), overcoming the shortcomings of current story visualization datasets, which are smaller and lower quality. Therefore, our method outperforms StoryGen in aesthetics and can be applied to the open-domain where the subjects and styles are considerably diverse.
We also conduct the qualitative comparisons with concurrent work TheaterGen~\cite{theatergen} on their benchmark, as shown in Fig.~\ref{fig:TheaterGen}. Two styles (anime style and real style) are also applied to our DreamStory to show its performance. As we can see from Fig.~\ref{fig:TheaterGen}, our DreamStory can generate more aesthetically pleasing images while maintaining better consistency across multiple subjects, such as the clothes of the boy and girl in the figure.
We present subject and scene images generated by our DreamStory with different seeds under both real-style and anime-style conditions, as illustrated in Fig.~\ref{fig:diversity}.
These extensive experimental results also prove the advantages of our DreamStory.

\subsection{Ablation Studies} \label{subsec:exp_ablation}

We also conduct ablation studies in our benchmark to verify the effectiveness of each of our modules in MSD. We integrated these two modules, Masked Mutual Self-Attention (MMSA) and Masked Mutual Cross-Attention (MMCA), individually into the baseline, \ie, Playground. All the settings are compared from both subjective and objective perspectives as described in Sec.~\ref{subsec:exp_metrics}.  

\subsubsection{Objective Comparison}
All the objective results are presented in Tab.~\ref{tab:results_ablation_benchmark}. It is evident from Tab.~\ref{tab:results_ablation_benchmark} that adding MMSA and MMCA improved subject consistency, as indicated by an increase in DreamSim similarity (DS) and D\&C-DS. However, a minor decline was observed in the similarity between the scene and its text. 
Quantitative analysis reveals that the CLIP-T score degradation primarily stems from background regions. The average CLIP-T score of background regions (masked subjects with 0/255 values) showed markedly greater reduction compared to complete scenes. 
Specifically, in the 3-subject benchmark, the background CLIP-T score drops by -0.0100, while the full scene declines only by -0.0050. A similar trend holds for the 2-subject benchmark, where the background decreases by -0.0051, compared to -0.0039 for the full scene.
This phenomenon is attributed to our generation process’s emphasis on subject consistency, which marginally affects the scene’s content. Nonetheless, this impact is negligible, \ie, the difference in CLIP-T similarity is less than 0.007. These results confirm that our method maintains subject consistency while preserving the scene’s semantics. Moreover, Tab.~\ref{tab:results_ablation_benchmark} reveals that incorporating both MMSA and MMCA modules led to our DreamStory achieving optimal performance regarding aesthetic scores and subject consistency. This conclusively validates the effectiveness of our approach.

\subsubsection{Subjective Comparison} 

We present the user study result of our DreamStory compared to our different settings, baseline, with MMCA and with MMSA in Fig.~\ref{fig:user_study}(d)(e). It can be seen from Fig.~\ref{fig:user_study} that the evaluators prefer DreamStory to the other settings. 
We also show the visual results of ablation studies in Fig.~\ref{fig:benchmark_ablation} to show the effectiveness of each component. As Fig.~\ref{fig:benchmark_ablation} illustrates, the approach without the MMSA module has the potential to generate images with blending subjects, particularly when two subjects are close within the image, as seen in the first row with the man and woman, and the second row with the pirate and boy. 
Furthermore, without the MMCA module, there is a significant discrepancy in appearance between the generated subject’s portraits and the reference image, as demonstrated in the second row’s pirate and boy and the fourth row’s boy and man. 
This discrepancy can be attributed to two factors. Firstly, the subject’s text contains rich appearance information about the target subject, which is aligned in the semantic space of the diffusion model. This alignment comes from the fact that the subject's reference portrait is generated by the same diffusion model using this text. Secondly, the lack of detailed text descriptions can lead to a substantial difference between the subject and the reference image during the generation process of the scene image.  This discrepancy can exacerbate the problem during the self-attention computation, as it hinders the identification of the correct patches when calculating patch similarity, resulting in a significant difference in appearance. 
Finally, our DreamStory achieves the best aesthetic and subject similarity performance by including both modules. This strongly validates the effectiveness of our method.

We conducted an ablation study to evaluate the necessity of (1) LLM rewriting and (2) accurate mask generation. 
As illustrated in Fig.~\ref{fig:LLM_rewriting}, without rewriting, the diffusion model fails to understand the names, such as \textit{Kondo}. This leads to generating incorrect types of subjects, such as generating gorillas as humans, as shown in the second row of Fig.~\ref{fig:LLM_rewriting}. 
Similarly, without our accurate mask generation strategy, the initial mask fails to fully cover the boy when the generated boy exceeds the initial mask region. This leads to inconsistent color generation in the hair region, as highlighted with the red circle in Fig.~\ref{fig:ABL_mask_gen}.
These results clearly validate the effectiveness of our proposed rewriting and mask generation approach.

\begin{table}[]
\setlength{\abovecaptionskip}{2pt} 
    \centering
    \small \renewcommand{\arraystretch}{1.25}
    \caption{Quantitative results of LLM accuracy (\%) on benchmark}
    \label{tab:LLM_accuracy}

\resizebox{\linewidth}{!}{ 
    \begin{tabular}{lcccc} 
    \bottomrule[1.5pt] 
     & 0-Subject & 1-Subject & 2-Subject & 3-Subject \\ \hline 
    LLaMA3-1B~\cite{dubey2024llama3} & 100.00 & 98.89 & 87.44 & 80.13 \\ 
    LLaMA3-3B~\cite{dubey2024llama3} & 100.00 & 98.88 & 91.44 & 84.67 \\ 
    ChatGPT4~\cite{openai2023GPT4}   & 100.00 & 98.86 & 95.29 & 91.28 \\ 
    \toprule[1.5pt] 
    \end{tabular}
}
\vspace{-10pt}
\end{table}

\begin{table*}[t]
\setlength{\abovecaptionskip}{2pt} 
\setlength{\belowcaptionskip}{-0.1cm}  
    \centering
    \small \renewcommand{\arraystretch}{1.25}
    \caption{Comparison results of different segmentation models. 
    Similar performance across various segmentation models demonstrates the robustness and efficiency of our DreamStory.
    }
    \label{tab:SAM_compare}
\begin{tabular}{p{5.4cm}| c c c c | c c c c}
\bottomrule[1.5pt]
\multicolumn{1}{l|}{\multirow{2}{*}{}}             & \multicolumn{4}{c|}{\textbf{2-Subject}} & \multicolumn{4}{c}{\textbf{3-Subject}} \\
\multicolumn{1}{l|}{}                              & AES$\uparrow$ & CLIP-T$\uparrow$  & DS$\uparrow$  &  D\&C-DS$\uparrow$   & AES$\uparrow$ & CLIP-T$\uparrow$  & DS$\uparrow$  &  D\&C-DS$\uparrow$   \\ \hline
w/ Grounded-Light-HQSAM~\cite{sam_hq}              & 6.71 & 0.3774  & 0.6725 & 0.5406 & 6.82 & 0.3792 & 0.5962 & 0.2327 \\
w/ Grounded-MobileSAM~\cite{mobile_sam}            & 6.72 & 0.3789  & 0.6707 & 0.5386 & 6.83 & 0.3799 & 0.5929 & 0.2284 \\
w/ Grounded-SAM~\cite{GroundedSAM} (\textbf{Ours}) & 6.72 & 0.3779  & 0.6714 & 0.5444 & 6.81 & 0.3791 & 0.5965 & 0.2335 \\ 
\toprule[1.5pt]
\end{tabular}
\vspace{-15pt}
\end{table*}

\begin{figure}[t]
\setlength{\abovecaptionskip}{2pt} 
\setlength{\belowcaptionskip}{-0.1cm}  
\centering
    \includegraphics[width=1.0\linewidth]{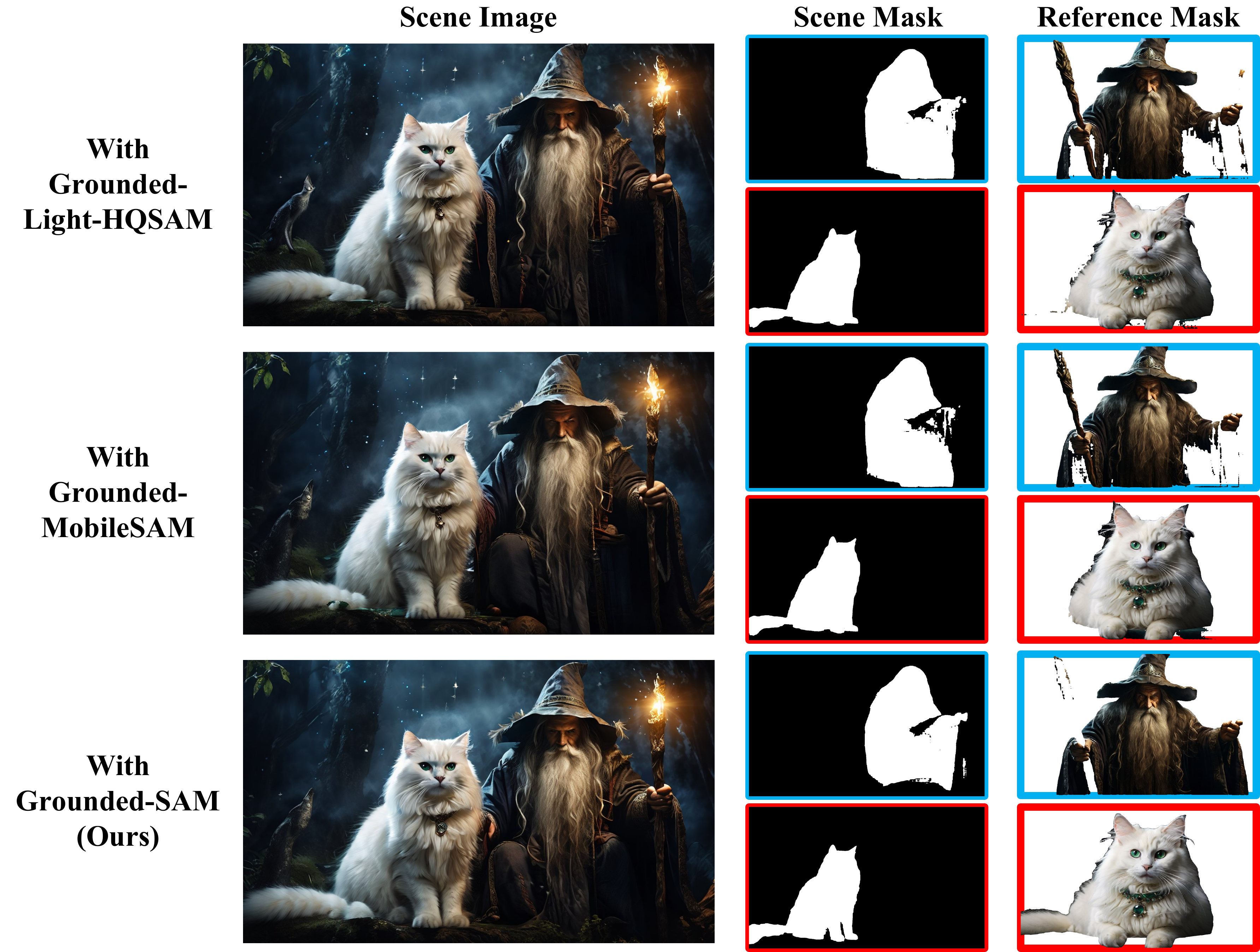}
    \caption{Qualitative comparisons of our DreamStory with different segmentation models. 
    Despite variations in masks produced by different SAM models, the final generated scene images remain nearly identical.
    This demonstrates the robustness of our method, showing its adaptability to different segmentation models.}
\label{fig:SAM_compare}
\vspace{-10pt}
\end{figure}

\subsection{Systematic Analysis} \label{subsec:sys_analyse}

In this subsection, we provide a comprehensive analysis to validate the performance and robustness of DreamStory. This includes a series of degradation tests and time-efficiency evaluations, ensuring a thorough assessment of our approach.

\subsubsection{Performance with Varying LLMs} \label{subsec:sys_analyse_LLM}
To assess the performance of DreamStory, we replaced two smaller LLMs (LLaMA3.2~\cite{dubey2024llama3}) and evaluated their accuracy in subject annotation. As shown in Tab.\ref{tab:LLM_accuracy}, all LLMs achieved over 98\% accuracy on the 0-Subject and 1-Subject benchmarks. For the 2-Subject and 3-Subject benchmarks, accuracy decreased slightly with smaller model sizes but remained above 80\%. Among the models tested, the largest LLM, ChatGPT4\cite{openai2023GPT4}, outperformed the others, achieving approximately 95\% accuracy for the 2-Subject and 91\% for the 3-Subject benchmarks.
Additionally, all LLMs successfully generated prompts for subjects and scenes that aligned with the story, likely due to their robust capabilities in text understanding and summarization—well-established tasks within LLM training data. ChatGPT, in particular, excelled at instruction-following, making it the model of choice for our approach. Fig.~\ref{fig:abl_llm} illustrates final story images generated using different LLMs, further emphasizing the robustness of DreamStory.

\subsubsection{Performance with Varying SAMs} \label{subsec:sys_analyse_SAM}

We also evaluated DreamStory's performance with different SAM models, specifically Light-HQSAM~\cite{sam_hq} and MobileSAM~\cite{mobile_sam}. The results, presented in Tab.\ref{tab:SAM_compare}, show that performance degradation with smaller SAMs across various metrics is minimal. Furthermore, Fig.\ref{fig:SAM_compare} displays example masks generated by different SAMs, along with their corresponding final scene images. While smaller SAMs occasionally miss fine details (such as the sleeves of wizard robes in the first two rows), the differences in the final generated images are negligible. These results affirm the robustness and high performance of DreamStory across varying SAM sizes.

\subsubsection{Time Efficiency Analysis} \label{subsec:sys_analyse_time}

The runtime of the DreamStory framework for generating a single story is primarily determined by the time required for the LLM to process requests, which typically takes 3 to 4 minutes, accounting for about 60\% of the total time. This time consumption is largely due to the waiting period for network-based API responses, which could be accelerated in future implementations.

In comparison, the image generation phase is faster, averaging 20 to 30 seconds per image. Additionally, the time required for SAM to generate masks is minimal, averaging less than 0.5 seconds, with negligible impact on the overall processing time.
Tab.~\ref{tab:time} summarizes the time required to generate a scene image using different methods. For finetuning-based approaches (e.g., MuDI), approximately 1.5 hours of finetuning is required for 2-subject scenarios, and 2 hours for 3-subject scenarios. In contrast, DreamStory generates each scene image in about 25 seconds, similar to other training-free approaches, highlighting its exceptional time efficiency.

\begin{table}[t]
\setlength{\abovecaptionskip}{2pt} 
\setlength{\belowcaptionskip}{-2pt}  
    \centering
    \small \renewcommand{\arraystretch}{1.25}
    \caption{Average time(s) Consumption on different methods. 
    It includes the time for generating reference subjects and the final scene. Additionally, MUDI requires extra fine-tuning time for each case.}
    \label{tab:time}
\begin{tabular}{p{3cm} p{2cm}<{\centering} p{2cm}<{\centering}}
\bottomrule[1.5pt]
                 & 2-Subject & 3-Subject \\
\hline
MuDI~\cite{MUDI}                        & 5400 & 7200 \\
ConsiStory~\cite{tewel2024_ConsiStory}  &  30  &  38  \\
StoryDiffusion~\cite{storydiffusion}    &  21  &  25  \\
\textbf{DreamStory}(\textbf{Ours})      &  22  &  28  \\
\toprule[1.5pt]
\end{tabular}
\vspace{-15pt}
\end{table}

\subsection{Limitations and Failure Cases} \label{subsec:exp_limitation}

\begin{figure}[t]
\setlength{\abovecaptionskip}{2pt} 
\centering
\includegraphics[width=1.0\linewidth]{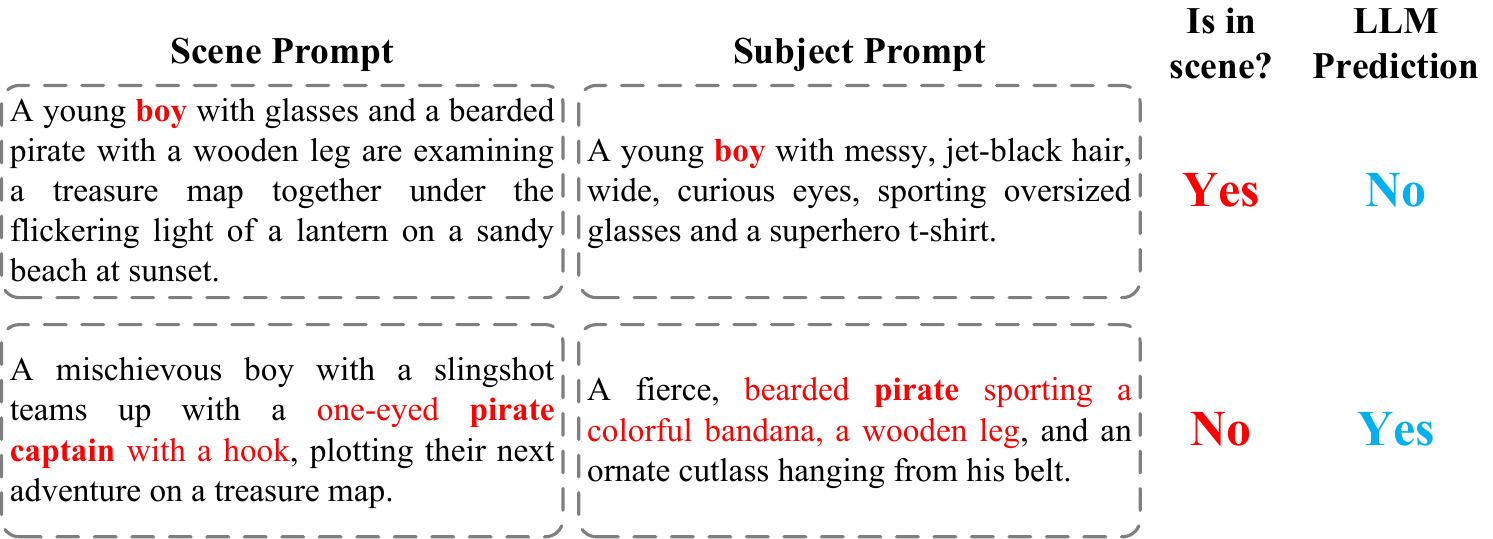}
\caption{Failure Cases of LLM.}
\label{fig:Failure_Cases_LLM}
\vspace{-10pt}
\end{figure}

\begin{figure}[t]
\setlength{\abovecaptionskip}{2pt} 
\setlength{\belowcaptionskip}{-0.1cm} 
\centering
\includegraphics[width=1.0\linewidth]{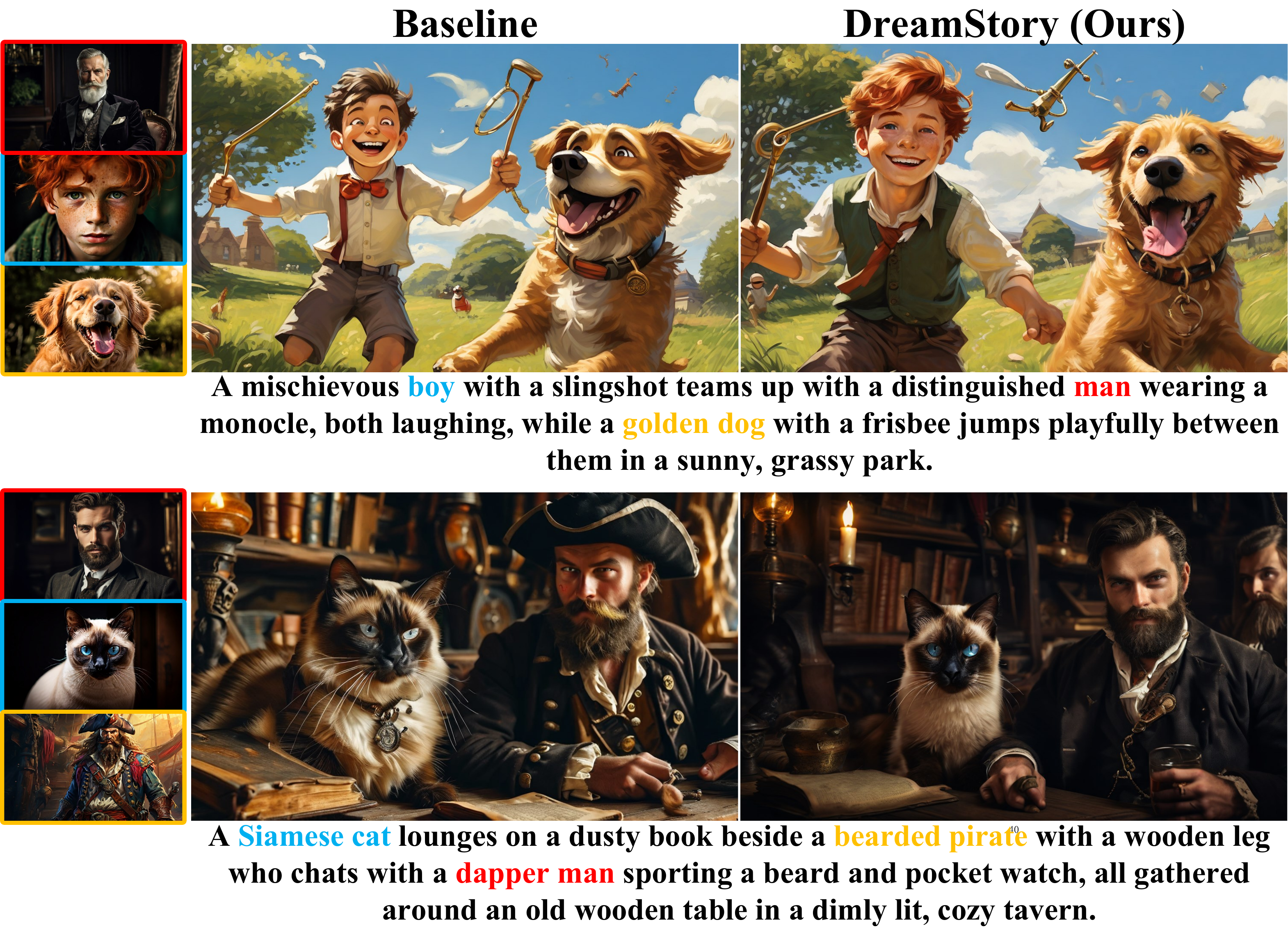}
\caption{Failure Cases of Diffusion Model. Different subjects are indicated with different colors.}
\label{fig:Failure_Cases_DM}
\vspace{-10pt}
\end{figure}

Our method relies on the abilities of both the LLM and diffusion model. Firstly, LLM may have hallucinations when labeling whether the subject is in the scene. As shown in Fig.~\ref{fig:Failure_Cases_LLM}, the scene includes a boy, but the LLM failed to identify the boy. In addition, the LLM may not effectively distinguish between subjects with similar descriptions, such as the \textit{pirate} and \textit{pirate captain} in the second row of Fig.~\ref{fig:Failure_Cases_LLM}.
Finally, diffusion models suffer from semantic understanding issues, which may be difficult in multi-subject and multi-attribute generation~\cite{MUDI}. As shown in Fig.~\ref{fig:Failure_Cases_DM}, when the diffusion model failed to generate three subjects for the first time, our method also failed to generate three subjects with consistent appearances.
Despite these limitations, our framework still has promising potential as individual models evolve and progress.

\section{Conclusion} \label{sec:conclusion}

This paper introduced an automatic training-free open-domain story visualization framework, DreamStory. It leverages Language Models (LLMs) as a story director to generate concise prompts for subjects and scenes, annotating the subjects in each scene. This information guides diffusion models in creating visually consistent content that aligns with the story narrative. We also developed a novel Multi-Subject consistent Diffusion model (MSD) that leverages both the subject prompt and its corresponding portrait to maintain consistency in multiple subjects across frames. To validate our approach and promote progress in story visualization, we established an evaluation benchmark, DS-500. Our method outperforms previous methods in aesthetics, image-text alignment, and subject consistency through objective and subjective evaluations.

In conclusion, our DreamStory method represents a significant step forward as a framework for open-domain story visualization. It does not require additional training and is poised to enhance its performance as the underlying models evolve. This positions our framework for promising advancements in story visualization.

\section*{Acknowledgments}
This work is supported by the National Natural Science Foundation of China (U1911203, U2001211, U22B2060), Guangdong Basic and Applied Basic Research Foundation (2019B1515130001), Key-Area Research and Development Program of Guangdong Province (2020B0101100001, 2024B0101050005), Research Foundation of Science and Technology Plan Project of  Guangzhou City (2023B01J0001, 2024B01W0004).

\normalem
\bibliography{ref}
\bibliographystyle{IEEEtran}
 
\vfill

\end{document}